\documentclass[pdflatex,sn-mathphys-num]{sn-jnl}

\usepackage{graphicx}
\usepackage{multirow}
\usepackage{amsmath,amssymb,amsfonts}
\usepackage{amsthm}
\usepackage{mathrsfs}
\usepackage[title]{appendix}
\usepackage{xcolor}
\usepackage{textcomp}
\usepackage{manyfoot}
\usepackage{booktabs}
\usepackage{algorithm}
\usepackage{algorithmicx}
\usepackage{algpseudocode}
\usepackage{listings}
\usepackage{enumitem}
\usepackage{pgfplots}
\usepgfplotslibrary{groupplots}
\usetikzlibrary{arrows.meta,positioning,fit}
\pgfplotsset{compat=1.17}
\definecolor{itoBlue}{HTML}{0072B2}
\definecolor{itoVerm}{HTML}{D55E00}
\definecolor{itoSky}{HTML}{56B4E9}
\definecolor{itoGreen}{HTML}{009E73}
\definecolor{itoOrange}{HTML}{E69F00}
\definecolor{itoPurple}{HTML}{CC79A7}
\definecolor{itoGray}{HTML}{666666}

\newcommand{\figontology}{%
\begin{tikzpicture}[
  font=\scriptsize,
  box/.style={draw, rounded corners=2.5pt, align=center, inner sep=3.2pt,
              fill=white, minimum height=0.74cm, minimum width=2.28cm},
  pay/.style={box, fill=itoGreen!14, draw=itoGreen!80!black},
  act/.style={box, fill=itoOrange!22, draw=itoOrange!80!black},
  inst/.style={box, fill=itoSky!18, draw=itoBlue, minimum width=2.02cm},
  arr/.style={-{Stealth}, thick, itoGray},
  darr/.style={-{Stealth}, densely dashed, thick, itoBlue},
]
\node[box] (user) at (0,2.75) {\textbf{User T2}};
\node[inst] (verb) at (-5.05,1.45) {verbalized conf.\\[-1pt]{\tiny Qwen2.5-7B}};
\node[act] (speech) at (-2.52,1.45) {speech act\\[-1pt]{\tiny re-ask / agree / accuse}};
\node[pay] (pay) at (2.52,1.45) {payload\\[-1pt]{\tiny gold / foil string}};
\node[inst] (ll) at (5.05,1.45) {likelihood};
\node[act] (ent) at (-2.52,0.10) {no directed gold\\[-1pt]{\tiny $\approx$0 I$\rightarrow$C}};
\node[pay] (ans) at (2.52,0.10) {answer update\\[-1pt]{\tiny I$\rightarrow$C / C$\rightarrow$I}};
\node[inst] (ptrue) at (5.05,0.10) {$P(\mathrm{True})$};
\node[pay] (com) at (2.52,-1.25) {commitment $s_t$\\[-1pt]{\tiny frozen $v_{\mathrm{acc}}$, Eq.~\eqref{eq:commit}}};
\node[inst] (probe) at (5.05,-1.25) {frozen probe};
\draw[arr] (user) -- (speech);
\draw[arr] (user) -- (pay);
\draw[arr] (speech) -- (ent);
\draw[arr] (pay) -- (ans);
\draw[arr] (ans) -- (com);
\draw[darr] (speech) -- (verb);
\draw[darr] (user.east) to[out=18,in=160] (ll.west);
\draw[darr] (ans) -- (ptrue);
\draw[darr] (com) -- (probe);
\end{tikzpicture}}

\newcommand{\figpayload}{%
\begin{tikzpicture}
\begin{axis}[
  ybar,
  bar width=4.4pt,
  width=\textwidth,
  height=4.0cm,
  ymin=0, ymax=85,
  ylabel={count / 200},
  symbolic x coords={persist,confirm,confirm+,empty,gold pl.,correct,foil pl.,persuade},
  xtick=data,
  x tick label style={rotate=18, anchor=east, font=\scriptsize},
  legend style={draw=none, font=\scriptsize, at={(0.02,0.98)}, anchor=north west},
  enlarge x limits=0.08,
  axis lines=left,
]
\addplot[fill=itoBlue, draw=none] coordinates {
  (persist,0) (confirm,2) (confirm+,0) (empty,12)
  (gold pl.,57) (correct,65) (foil pl.,1) (persuade,0)};
\addplot[fill=itoVerm, draw=none] coordinates {
  (persist,5) (confirm,8) (confirm+,2) (empty,71)
  (gold pl.,10) (correct,15) (foil pl.,77) (persuade,30)};
\legend{I$\rightarrow$C, C$\rightarrow$I}
\end{axis}
\end{tikzpicture}}

\newcommand{\figinstruments}{%
\begin{tikzpicture}
\begin{axis}[
  ybar,
  bar width=5.2pt,
  width=7.1cm,
  height=3.1cm,
  ymin=-0.32, ymax=0.22,
  ylabel={mean $\Delta$ (T2$-$T1)},
  symbolic x coords={persist,confirm,persuade,correct},
  xtick=data,
  x tick label style={font=\scriptsize},
  legend style={draw=none, font=\scriptsize, at={(0.0,-0.02)}, anchor=north west, legend columns=2},
  enlarge x limits=0.18,
  axis lines=left,
]
\addplot[fill=itoOrange, draw=none] coordinates {(persist,-0.104) (confirm,-0.102) (persuade,-0.173) (correct,-0.150)};
\addplot[fill=itoSky, draw=none] coordinates {(persist,0.140) (confirm,0.142) (persuade,0.020) (correct,0.088)};
\addplot[fill=itoGreen, draw=none] coordinates {(persist,0.153) (confirm,0.153) (persuade,-0.034) (correct,0.166)};
\addplot[fill=itoPurple, draw=none] coordinates {(persist,0.007) (confirm,0.002) (persuade,-0.258) (correct,-0.003)};
\legend{$\Delta$verb, $\Delta$ll, $\Delta$probe, $\Delta P$(True)}
\end{axis}
\end{tikzpicture}}

\begin{document}

\title[Not One Belief]{Not One Belief: Answer Payloads, Persistence, and Dissociated Confidence Signals in Conversational Question Answering}

\author*[1]{\fnm{Duc-Vinh} \sur{Tran}}\email{vinh.td225538@sis.hust.edu.vn}

\affil*[1]{\orgdiv{School of Information and Communication Technology}, \orgname{Hanoi University of Science and Technology}, \orgaddress{\city{Hanoi}, \postcode{100000}, \country{Vietnam}}}

\abstract{Conversational question-answering systems are expected to fix their errors when users push back, and to signal which answers to trust. In short-answer question answering, we split each user follow-up into its speech act (re-ask, agree, or say the answer is wrong) and its payload (an answer string, correct or not). On the same transcripts we measure four confidence signals: verbalized confidence, length-normalized sequence likelihood, a hidden-state linear probe fitted at the first turn and then frozen, and the self-evaluation probability P(True). Answers change when the user supplies an answer string. They almost never change under the speech act alone: in every retained full-size model--dataset cell, the correct answer converts at least four times as many wrong answers as re-asking or agreement. The four signals disagree on the same turns. On Qwen2.5-7B the frozen probe rises both on answers that stay wrong and on answers that switch to the correct one, so it goes with answer persistence rather than truth. This survives scoring on held-out questions, and Llama-3-8B with cleaned answers supports it; on four other models it does not hold up on held-out questions. For 81 Qwen2.5-7B errors that survive one re-ask, re-asking, a fresh chat, or an abstain option repairs almost none; the correct answer repairs 75--95\%. In two preregistered replications on four models, every observed repair rate without an answer string stays below a preset 0.15, and a post hoc rerun with about 300 such items per model excludes 0.15 for every arm except the empty elicit. With one BM25-retrieved TriviaQA sentence, a sentence that contains the answer repairs 67--76\% of persistent errors, and one that lacks it repairs at most 12\% (16\% under model-read labels). In a post hoc end-to-end test with BM25 passages, gating evidence on whether it contains the gold answer cut harm on TriviaQA, but a gate that does not know the answer repaired few errors, so the check remains a heuristic.}

\keywords{Large language models, Confidence calibration, Multi-turn dialogue, Sycophancy, Probing, Retrieval-augmented generation}

\maketitle

\section{Introduction}\label{sec:intro}

Chat assistants built on large language models (LLMs) now answer questions in customer support, tutoring, search, and clinical or legal help lines. There, a wrong answer costs something. Two questions follow, both about the reliability of conversational QA and evidence retrieval under user feedback. When a user pushes back, does the system fix its errors or just change its mind? And can the system's own confidence tell the operator which answers to trust? Recent work on the second question asks whether LLM confidence stays calibrated as a conversation goes on~\cite{zhang-etal-2026-confidence-estimation,zhang-etal-2026-mtcal}. These studies usually score one confidence measure (verbalized confidence, the self-evaluation probability $P(\mathrm{True})$, or sequence likelihood) against correctness, as if each were an estimate of the same underlying belief.

\begin{figure*}[t]
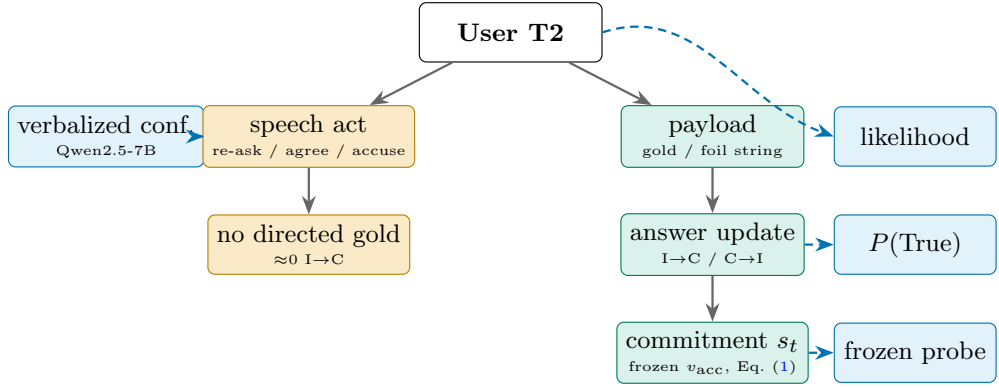

\centering
\resizebox{\textwidth}{!}{\figontology}
\caption{Which part of the second user turn (T2) changes which quantity. Solid lines show what a controlled T2 message changes; dashed lines show which confidence measure attaches to it. Green boxes form the payload chain, orange boxes the speech-act chain, and blue boxes are the confidence measures. The persistence readout $s_t$ is defined in Section~\ref{sec:commit-def}}
\label{fig:ontology}
\end{figure*}

For that to work, the measures must move together, and every follow-up must count as the same kind of evidence. In our data, neither holds. So we ask a simpler question: \emph{when a user sends a follow-up message, which quantity does it change?} Figure~\ref{fig:ontology} summarizes the answer. We split each follow-up into two parts. The \emph{speech act} is what the user does: re-ask, agree, or say the model is wrong. The \emph{payload} is an optional answer string in the message, either the correct (``gold'') answer or a plausible wrong one (a ``foil''). Payloads change the model's answer. Speech acts alone almost never repair an error. Changes to the answer, whether or not the new answer is true, move a frozen probe readout that, on Qwen2.5-7B, goes with answer persistence. We call it the \emph{persistence readout} $s_t$ and make no psychological or causal claim about it. The other measures respond to other things: verbalized confidence to the number of turns, sequence likelihood to how fluently the model repeats itself, and $P(\mathrm{True})$ to explicit contradiction. Two papers that calibrate different measures and call both ``confidence'' are not reporting comparable numbers.

Recent single-turn work already shows that confidence measures can come apart. \citet{kumaran2026commitment} finds that verbal confidence predicts a model's later decision to commit to or withhold an answer better than it predicts correctness, while calibrated token log-probabilities track correctness. \citet{zhang-etal-2026-confidence-estimation} study multi-turn confidence as hints accumulate and propose InfoECE and $P(\mathrm{Sufficient})$, but they do not separate what the user does from the answer string the user gives. \citet{kumaran2026commitment} work in a single turn. Our contribution is to control the user's turn in multi-turn dialogue, separate the speech act from the answer payload, and measure four signals on the same transcripts.

\paragraph{What moves the answer.}
We write I$\to$C for an item that is incorrect at turn~1 and correct at turn~2, and C$\to$I for the reverse. Re-asking (\texttt{persist}) and agreeing with the model (\texttt{confirm}) produce almost no I$\to$C changes, so the error survives. Stating the gold string produces many, with or without ``you are wrong'' around it. Stating a plausible foil produces many C$\to$I changes. An empty challenge (``I think that's wrong'') makes answers unstable without pointing them anywhere. The string moves the answer. The stance does not. Models differ only in how they take up the string. Qwen2.5 and Llama-3-8B usually copy it verbatim (the fraction of I$\to$C answers that copy it is $0.70$--$0.91$ on NQ and HotpotQA). Mistral-7B-v0.3 restates it inside a longer sentence.

\paragraph{The measures disagree.}
On Qwen2.5-7B-Instruct, verbalized confidence falls as the user keeps talking, even when accuracy is flat (\texttt{persist}) or rising (\texttt{correct}). Length-normalized sequence likelihood often rises on the same turns. A linear probe on hidden states, fitted once on first-turn answers and then frozen, rises both on items that stay wrong and on items that switch to gold. Kadavath-style $P(\mathrm{True})$~\cite{kadavath2022mostly} barely moves, except under foil persuasion, where it collapses on C$\to$I changes (mean change $\approx-0.26$ overall and $-0.85$ on C$\to$I items). One conversation, four different stories. The same holds for calibrated scores: a single user turn sends the expected calibration error (ECE) down for verbalized confidence, up for likelihood, and nowhere for $P(\mathrm{True})$ (Section~\ref{sec:inst}).

\paragraph{Persistent errors are hard to undo.}
Asking again or starting a new chat does not fix a persistent error unless the user supplies a string. Among 81 Qwen2.5-7B TriviaQA items that are still wrong after one re-ask, re-asking through turn~5, a fresh chat, and an \texttt{ABSTAIN} option repair two, zero, and zero items. The model abstains on $94\%$ of these stay-wrong items and on $62\%$ of stay-correct items. A gold string at turn~3 repairs $0.75$--$0.95$ of the same items, depending on phrasing, and an ``I don't know'' turn repairs $5/81$. Two preregistered replications (230 stay-wrong items from a second Qwen slice, 160 from Llama-3-8B, 139 from Mistral-7B-v0.3, and 61 from Qwen2.5-32B) keep every observed repair rate without an answer string below a $0.15$ threshold fixed in advance, under the preregistered answer matcher. A post hoc rerun on about 300 new stay-wrong items per model excludes $0.15$ for every such arm except the empty elicit (Section~\ref{sec:l2}). With one BM25 sentence per TriviaQA question, retrieval repairs errors only to the extent that the sentence contains the answer string (Section~\ref{sec:l2}).

\paragraph{Contributions.}
We offer a map of interventions and their effects, not a new method:
\begin{enumerate}[leftmargin=1.5em]
\item \textbf{The answer string, not the speech act, moves the answer.} Answers follow a stated gold or foil string. Re-asking and agreement do not repair errors (Section~\ref{sec:payload}).
\item \textbf{The measures come apart across turns.} Single-turn work has shown that verbalized confidence and correctness-tracking signals are not interchangeable~\cite{kumaran2026commitment,miao2026confidence}. We show the same split in dialogue for four measures (verbalized confidence, likelihood, a frozen hidden-state probe, and $P(\mathrm{True})$), in raw scores and in calibration error, and we show which part of the user turn each one responds to (Section~\ref{sec:inst}).
\item \textbf{On Qwen2.5-7B the frozen probe goes with answer persistence, and persistent errors are hard to undo.} The persistence readout $s_t=\sigma(v_{\mathrm{acc}}^{\top}\tilde{h}_t)$ is a frozen probe on the current answer state (Section~\ref{sec:commit-def}). On Qwen2.5-7B it rises on stay-wrong items \emph{and} on I$\to$C items, also on held-out questions and above a shuffled-label control. Llama-3-8B with cut answers supports this; on the other four models it does not hold on held-out questions. Resetting the chat or offering abstention does not recover these errors; a gold string at turn~3 does (Section~\ref{sec:l2}).
\item \textbf{A retrieval heuristic.} Before relying on evidence to repair an answer, check that it contains the answer string. This is a hypothesis for system design, not a validated rule. Across four models, a retrieved sentence without it repairs at most $0.16$ of stay-wrong items under any of our scorers, and it breaks up to $0.70$ of correct answers (Section~\ref{sec:l2}).
\end{enumerate}
The first three points share one argument: \emph{if two measures respond to different controlled interventions and disagree on the same transcript, they cannot both estimate a single latent conversational confidence.} We do not propose a new uncertainty estimator. Section~\ref{sec:limits} lists what we do not claim.

\section{Related work}\label{sec:related}

\paragraph{Sycophancy.}
Instruction-tuned LLMs tend to adopt a view the user states, even when it is wrong. \citet{perez2022sycophancy} documented this with model-written evaluations and found that it grows with model size and with reinforcement learning from human feedback. \citet{sharma2023sycophancy} showed that sycophancy is common across production assistants and traced it partly to human preference judgments that reward agreement. \citet{wei2023simple} found that scaling and instruction tuning increase sycophancy, that models agree with a user's clearly false arithmetic claims, and that fine-tuning on simple synthetic data reduces it. These works show that a user's stated answer is persuasive. They usually bundle the stance (``I think you are wrong'') with the stated string. Our factorial design separates the two. For short factual answers, the string does almost all of the work.

\paragraph{Re-asking and answer flipping under pushback.}
Other work studies what happens when a user challenges an answer, with or without offering a new one. In the FlipFlop experiment, \citet{laban2023flipflop} showed that a simple ``Are you sure?'' makes models change their answers and lowers accuracy. \citet{xie2024askagain} found that models waver under skeptical, negating, or misleading follow-ups even when the first answer was correct, and proposed a training method to reduce this. \citet{wang2023defend} showed that ChatGPT often cannot defend a correct answer against invalid arguments. \citet{huang2024selfcorrect} found that models rarely improve their reasoning by intrinsic self-correction without outside feedback, and \citet{laban2025lost} showed that performance drops across many tasks in multi-turn conversation. Our results agree on one side and add the other: empty challenges break correct answers (C$\to$I), but neither challenges nor re-asks repair wrong ones (I$\to$C). We also test whether a persistent error survives a chat reset.

\paragraph{Verbalized confidence.}
Asking a model to state its confidence is the cheapest measure in deployed systems. \citet{lin2022teaching} showed that a fine-tuned GPT-3 can express calibrated uncertainty in words. Without such fine-tuning, the stated confidence of instruction-tuned models is often overconfident or poorly calibrated~\cite{xiong2024verbalized}, though prompting can help, especially for models trained with human feedback~\cite{tian2023just}. \citet{geng2024survey} survey confidence estimation and calibration for LLMs. Calibration is usually measured with ECE~\cite{guo2017calibration}. In our dialogues, the movement of verbalized confidence is mostly a Qwen2.5-7B effect of turn depth, not a response to evidence.

\paragraph{Self-evaluation with $P(\mathrm{True})$.}
\citet{kadavath2022mostly} asked a model whether its own proposed answer is true and read the probability of ``True''. They found this fairly well calibrated for large models on many single-turn tasks. In our dialogues $P(\mathrm{True})$ ranks correct answers well at both turns but barely moves under re-asking or agreement. It responds mainly to contradiction by a foil.

\paragraph{Probing hidden states for truthfulness.}
Several papers train classifiers on hidden activations to read out whether a statement or answer is true. \citet{azaria2023internal} trained such a classifier on true and false statements. \citet{marks2023geometry} reported linear structure that separates true from false statements. \citet{burns2023ccs} found latent knowledge without labels. \citet{orgad2025llmsknow} showed that truthfulness signals concentrate in the exact answer tokens, generalize poorly across datasets, and that a model may encode the correct answer while generating a wrong one. A probe can learn the task itself rather than reveal what the representation encodes, so probe results need care~\cite{hewitt2019control}. We fit a probe at turn~1, freeze it, and follow it after the user speaks again.

\paragraph{Confidence as commitment.}
Two recent single-turn studies are close to our result on the measures. \citet{kumaran2026commitment} used a two-stage task: a model answers and reports its confidence, then later decides whether to give the answer to a user or abstain. Across several models, verbal confidence predicted the later commit-or-abstain decision better than it predicted correctness, while calibrated token log-probabilities tracked correctness. In Gemma models, a post-answer hidden state encoded the later decision, and steering along a direction specific to verbal confidence shifted abstention. \citet{miao2026confidence} found that a linear accuracy probe and the direction behind verbalized confidence are nearly orthogonal, and used activation steering to bring verbalized confidence closer to the probe. We do not claim these points as new. Our study differs in three ways. The user turn is controlled, so we can separate the speech act from the answer string. The probe is fitted at turn~1 and frozen, and we follow the same items across turns. And we study the frozen probe readout, not verbalized confidence: in our dialogues, on Qwen2.5-7B, it is the probe that moves with the answer the model keeps. We run no steering, so unlike these two studies our evidence is correlational. \citet{sarkar2026contextual} show, in a collaborative dialogue task between two LLMs, that a partner's assertion can shift a model's linear representation of whether a proposition is true. Our probe result points the same way for short factual answers.

\paragraph{Multi-turn calibration.}
The closest work studies confidence across turns. \citet{zhang-etal-2026-confidence-estimation} evaluate verbalized confidence, $P(\mathrm{True})$, and a new $P(\mathrm{Sufficient})$ score as hints accumulate, and propose an information-aware ECE (InfoECE) that requires confidence to rise steadily as gold information arrives. \citet{zhang-etal-2026-mtcal} show that persuasion worsens a turn-level ECE (ECE@T) and that sequence likelihood can \emph{rise} when the answer flips to a wrong one, and they train a calibrator, MTCal, to repair the score. Both ask how well a score tracks correctness. We use their setting (short-answer dialogue under persuasion) and ask an earlier question: which quantity a given user turn moves (Table~\ref{tab:related}).

\paragraph{Knowledge conflicts and retrieval.}
Models follow context that contradicts their stored knowledge to varying degrees~\cite{longpre2021entity,xie2024chameleon}, and whether retrieved context is sufficient to answer the question predicts errors in retrieval-augmented generation~\cite{joren2025sufficient}. \citet{lathkar2026anchored} reports that, in multi-hop retrieval-augmented QA, one confirmed intermediate fact can raise the rate of confident wrong answers before complete evidence lowers it. That work is single-turn and measures confidence; we measure whether evidence repairs an error that has already survived a re-ask. In our data, evidence repairs such an error only when it contains the answer string.

\begin{table}[t]
\centering
\caption{How this work relates to the closest prior work. ``Left open'' lists what each design does not separate}
\label{tab:related}
\small
\begin{tabular}{p{0.20\textwidth}p{0.34\textwidth}p{0.36\textwidth}}
\toprule
 & Question asked & Left open \\
\midrule
\citet{zhang-etal-2026-confidence-estimation} & Estimation as evidence accumulates; InfoECE; $P(\mathrm{Sufficient})$ & Frozen hidden-state readout; gold vs.\ foil vs.\ empty challenge; whether verbalized confidence, likelihood, and $P(\mathrm{True})$ agree on one transcript \\
\citet{zhang-etal-2026-mtcal} & ECE@T after persuasion; likelihood; a learned calibrator & Verbalized confidence vs.\ frozen probe; agreement and correction turns; reset; removing the answer string \\
\citet{kumaran2026commitment} & Does verbal confidence track correctness or a later commit/abstain decision? (single turn) & User follow-up turns; behaviour across turns \\
\citet{miao2026confidence} & Accuracy probe vs.\ verbalized-confidence direction; steering (single turn) & User follow-up turns; likelihood and $P(\mathrm{True})$ \\
Sycophancy~\cite{perez2022sycophancy,sharma2023sycophancy} & User states $X$ $\Rightarrow$ model says $X$ & Separating ``you are wrong'' from the stated string \\
This work & Which user intervention changes which quantity across turns; $s_t$ is not truth & A new estimator; causal tests such as steering \\
\bottomrule
\end{tabular}
\end{table}

\section{Experimental setup}\label{sec:setup}

\subsection{Dialogue protocol}
Each dialogue starts at turn~1 (T1) with a short-answer question. At turn~2 (T2), and at later turns when used, the user sends one of four \emph{core} messages, fixed across models:
\begin{description}
\item[\texttt{persist}] \emph{Please answer again, briefly. Do not change your answer unless you are sure you were wrong.}
\item[\texttt{confirm}] \emph{Yes, that's correct. Please answer again, briefly, in one short phrase.}
\item[\texttt{persuade}] One of five challenge templates plus a foil.
\item[\texttt{correct}] The same five templates plus a gold alias.
\end{description}
Ablations keep the same T1 and change only the T2 message, to separate the speech act from the payload. \texttt{gold\_plain} and \texttt{foil\_plain} state a string without saying ``wrong''. \texttt{challenge\_empty} challenges the answer without offering a replacement. \texttt{confirm\_strong} repeats the model's T1 answer and calls it correct. \texttt{persist\_short} is \emph{Again, briefly.} Exact wording is in Online Resource~1, Sect.~A.

\subsection{Notation and outcome categories}
We compare each item's correctness at turn $t$ with its correctness at T1. An item is \emph{stay-correct} (SC) if it is correct at both turns, \emph{stay-wrong} (SW) if it is wrong at both, I$\to$C if it changes from incorrect to correct, and C$\to$I if it changes from correct to incorrect. We report I$\to$C and C$\to$I as counts out of the number of items (usually 200) or as rates. For a measure $X$, $\Delta X = X_{\mathrm{T2}} - X_{\mathrm{T1}}$ is its change from T1 to T2, averaged over the stated items. The \emph{copy rate} is the fraction of I$\to$C answers that repeat the payload string exactly. A \emph{unit} is one model$\times$dataset combination. Seed copies of the same unit are not counted as independent.

\subsection{Confidence measures}
We use the four quantities that current multi-turn papers treat as interchangeable measures of confidence. The list is not exhaustive. Token entropy, logit margin, and semantic entropy~\cite{kuhn2023semantic} read the same next-token distribution as likelihood, so we expect them to behave like it unless an intervention separates them.
\begin{itemize}
\item \textbf{Verbalized confidence} ($\Delta$verb for its change) is what products elicit~\cite{xiong2024verbalized,tian2023just}. The model is asked to state a confidence, which is parsed from a separate \texttt{CONF} field into $[0,1]$.
\item \textbf{Sequence likelihood} (``ll'') is the quantity scored by ECE@T and MTCal~\cite{zhang-etal-2026-mtcal}: $\exp$ of the mean log-probability of the answer tokens, teacher-forced under the chat template.
\item \textbf{Frozen probe} is the kind of hidden-state readout that the truthfulness-probing literature reads as a signal of truth~\cite{azaria2023internal,marks2023geometry}. It is defined in Section~\ref{sec:commit-def}.
\item \textbf{$P(\mathrm{True})$} is the Kadavath self-evaluation used alongside $P(\mathrm{Sufficient})$~\cite{kadavath2022mostly,zhang-etal-2026-confidence-estimation}: a forced choice between A (true) and B (false), scored by the softmax probability of A. We computed it on the Qwen2.5-7B TriviaQA set of 800 rows (200 items $\times$ four core messages).
\end{itemize}

\subsection{The persistence readout: an operational definition}\label{sec:commit-def}
Let $h_t\in\mathbb{R}^d$ be the last-layer hidden state averaged (mean-pooled) over the answer tokens at turn $t$, and let $\tilde{h}_t$ be $h_t$ standardized with the training-set mean and variance. We fit a weight vector $v_{\mathrm{acc}}\in\mathbb{R}^d$ by logistic regression to predict T1 correctness from $h_1$, using only the \emph{unique} \texttt{persist} T1 items of that model$\times$dataset. GroupKFold cross-validation is used only as a sanity check. We then \emph{freeze} $v_{\mathrm{acc}}$ and never refit it after the user speaks. The persistence readout at turn $t$ is
\begin{equation}
s_t \;=\; \sigma\!\bigl(v_{\mathrm{acc}}^{\top} \tilde{h}_t\bigr),
\label{eq:commit}
\end{equation}
where $\sigma(x)=1/(1+e^{-x})$ is the logistic function. We use ``probe'' and $s_t$ interchangeably and write $\Delta s=s_2-s_1$. Three points make the definition precise:
\begin{itemize}
\item \textbf{What it is.} $s_t$ is a frozen readout of the current answer state. The linear map is one choice. Our claims about $s_t$ apply to any frozen readout that is fitted only on unique T1 states, is not refitted after the user speaks, and rises on stay-wrong \emph{and} on I$\to$C items (Section~\ref{sec:commit}).
\item \textbf{What the name means.} ``Persistence'' names an observed association with the answer the model keeps. We make no claim about a psychological state, and no claim that $s_t$ causes the answer. The main analyses fit the probe on unique T1 items and apply it to rows of the same questions. A post hoc rerun on held-out questions (Section~\ref{sec:commit}) keeps the result on Qwen2.5-7B only.
\item \textbf{Why it is not truth.} If $s_t$ measured truth, it would rise only when the answer becomes gold. Instead it rises on stay-wrong and on I$\to$C items and falls on C$\to$I items.
\item \textbf{The behaviour it goes with.} An item is \emph{stay-wrong} if it is wrong at T1 and still wrong after \texttt{persist} at T2. On Qwen2.5-7B TriviaQA there are $n{=}81$ such items, and $\Delta s>0$ on 79 of their 80 distinct questions. Reset and abstention fail on this set; a gold string at T3 does not (Section~\ref{sec:l2}).
\end{itemize}
We do not read $v_{\mathrm{acc}}$ as a discovered truth direction~\cite{marks2023geometry,burns2023ccs}. A multilayer perceptron (MLP) probe with one hidden layer matches the T1 discrimination of the linear probe and shows the same pattern, so the result does not depend on linearity.

\subsection{Answer matching}
Answers are normalized (lowercased, articles removed, whitespace collapsed) and counted correct on an exact match or on whole-word containment, in either direction, between the answer and a gold alias~\cite{joshi2017triviaqa}. We call this the \emph{contain} matcher. It is the primary, preregistered scorer. Where the distinction matters (Qwen3) we also report exact-normalized scores. Containment has a known weakness: it credits an answer that names the gold only to reject it (\emph{Norman Rockwell, not Edward Hopper}). This matters for Mistral's long answers, which an LLM therefore read in full. Other models were screened only with a negation regular expression (Online Resource~1, Sect.~I). The second preregistration also scored every answer with an LLM judge. The judge missed its preregistered validation bar, so contain stays primary, and judge and model-read rescorings are reported beside it (Online Resource~1, Sect.~H). Post hoc, two independent people labelled a blind sample of $300$ V2 rows; on the $272$ clear ENDORSES/DOES\_NOT pairs Cohen's $\kappa$ was $0.86$ (agreement $253/272$), with disagreements resolved by discussion. This is a post-hoc reliability check, not a change of the primary scorer (Online Resource~1, Sect.~H).

Under our generation settings, Llama-3-8B-Instruct does not stop at its end-of-turn token \texttt{<|eot\_id|>} and repeats its answer after a leaked \texttt{assistant} role header. For Llama we score the text before the first leaked role header, because in the raw run-on text the matcher misses answers glued to the header ($9$--$19\%$ of Llama T2 labels change).

\subsection{Models and data}
The main grid uses instruction-tuned models from 3B to 14B parameters: Qwen2.5-\{3,7,14\}B~\cite{qwen25}, Qwen2-7B, Qwen3-8B~\cite{qwen3}, Llama-3-8B-Instruct~\cite{llama3}, Mistral-7B-Instruct-v0.3~\cite{mistral7b}, and Phi-4~\cite{phi4}. Qwen2.5-32B-Instruct has one TriviaQA T1/T2 cell. We use it only to find stay-wrong items for the turn-3 battery and the retrieval study (Section~\ref{sec:l2}), and it is not one of the 24 units below. We had no access to the gated Llama-3.1 and Gemma checkpoints.

Datasets are TriviaQA \texttt{rc.nocontext}~\cite{joshi2017triviaqa}, NQ-open~\cite{kwiatkowski2019nq}, HotpotQA questions without passages~\cite{yang2018hotpot}, TruthfulQA~\cite{lin2022truthfulqa}, and WebQuestions~\cite{berant2013webq}. For TriviaQA we use a main slice $[0,200)$, fixed after shuffling with seed~42, plus a disjoint slice $[200,400)$ and an extra slice of $n=500$. GSM8K~\cite{cobbe2021gsm8k} uses the same short-answer protocol on grade-school mathematics (matching the last number); we make no claim about long-form reasoning. We \emph{exclude} PopQA because several models fall to accuracy~0, a matcher and format failure. Unless stated otherwise, $n=200$, seed~42, and decoding is greedy.

\subsection{Replication predicates}\label{sec:predicates}
To check whether each pattern generalizes, we evaluate five yes/no predicates on every independent unit:
\begin{itemize}
\item $\mathrm{P}_{\mathrm{err}}$: on stay-wrong items, $\Delta$probe $>0$ under both \texttt{persist} and \texttt{confirm}.
\item $\mathrm{P}_{\mathrm{gold}}$: I$\to$C under \texttt{correct} is at least $0.15$, and I$\to$C is at most $0.03$, $0.05$, and $0.05$ under \texttt{persist}, \texttt{confirm}, and \texttt{persuade}.
\item $\mathrm{P}_{\mathrm{verb}}$: mean $\Delta$verb $<0$ on all four core messages (models whose verbalized confidence is flagged \texttt{VERB\_UNTRUSTED} are skipped).
\item $\mathrm{P}_{\mathrm{ll}}$: under \texttt{persist}, the most common sign combination is verbalized confidence not up and likelihood up, or mean $\Delta$ll $>0$ with mean $\Delta$verb $<0$.
\item $\mathrm{P}_{\mathrm{artifact}}$: at least $80\%$ of C$\to$I changes under \texttt{correct} are items whose T1 answer was already gold under a loose alias check, that is, a matcher artifact rather than a real loss.
\end{itemize}
On 24 independent units, $\mathrm{P}_{\mathrm{err}}=\mathrm{P}_{\mathrm{ll}}=\mathrm{P}_{\mathrm{artifact}}=24/24$. $\mathrm{P}_{\mathrm{gold}}$ holds on $19/24$. The five failures are the three Qwen3 units, which we keep in the denominator although their cells are truncated ``self-repair'' cells that we withdrew after extracting answers past \texttt{</think>} (Online Resource~1, Sect.~C), and Llama-3-8B on TriviaQA and NQ (Section~\ref{sec:payload}). $\mathrm{P}_{\mathrm{verb}}$ holds on all six Qwen2.5-7B dataset units and on no other model family.

\subsection{Preregistration and what it cost}\label{sec:prereg}
Before any generation, a frozen pilot preregistration~\cite{prereg-pilot} fixed two hypotheses and a stopping rule. H1 stated that the frozen $v_{\mathrm{acc}}$ predicts T2 correctness after misleading persuasion better than verbalized confidence does. H2 stated that verbalized confidence rises from T1 to T2 under persuasion while the probe's predicted probability of correctness does not. H2 failed its gate: verbalized confidence fell ($-0.17$). H1 met neither preregistered bar: the advantage was $+0.06$ with a bootstrap interval covering zero. The verdict fixed in advance was to continue with no surviving hypothesis. We report H2 as failed rather than dropping it. What replaced the hypotheses was an observation: the measures were responding to different parts of the user's turn.

Apart from two later preregistered studies of persistent errors and retrieval~\cite{prereg-t3rep,prereg-ext} (Section~\ref{sec:l2}), nothing in this paper is confirmatory. Every other result is exploratory measurement, and we label it as such. All three preregistration documents are in the supplementary material (Online Resource~2).

\section{The answer string, not the speech act}\label{sec:payload}

\begin{table}[t]
\centering
\caption{Core T2 messages on Qwen2.5-7B-Instruct, TriviaQA ($n=200$, seed~42). I$\to$C and C$\to$I are counts out of 200; ``acc T2'' is accuracy at T2; $\Delta$verb is the mean change in verbalized confidence. \texttt{correct\_long} drops the word ``briefly''; its I$\to$C stays at 67 (copy rate $0.60$)}
\label{tab:g6}
\small
\begin{tabular}{lrrrr}
\toprule
T2 & I$\to$C & C$\to$I & acc T2 & $\Delta$verb \\
\midrule
\texttt{persist} & 0 & 5 & .570 & $-0.10$ \\
\texttt{confirm} & 2 & 8 & .565 & $-0.10$ \\
\texttt{confirm\_strong} & 0 & 2 & .585 & $-0.06$ \\
\texttt{persist\_short} & 4 & 12 & .555 & $-0.01$ \\
\texttt{challenge\_empty} & 12 & 71 & .300 & $-0.16$ \\
\texttt{gold\_plain} & 57 & 10 & .830 & $-0.08$ \\
\texttt{correct} & 65 & 15 & .845 & $-0.15$ \\
\texttt{foil\_plain} & 1 & 77 & .215 & $-0.14$ \\
\texttt{persuade} & 0 & 30 & .445 & $-0.17$ \\
\texttt{correct\_long} & 67 & -- & -- & -- \\
\bottomrule
\end{tabular}
\end{table}

\begin{figure*}[t]
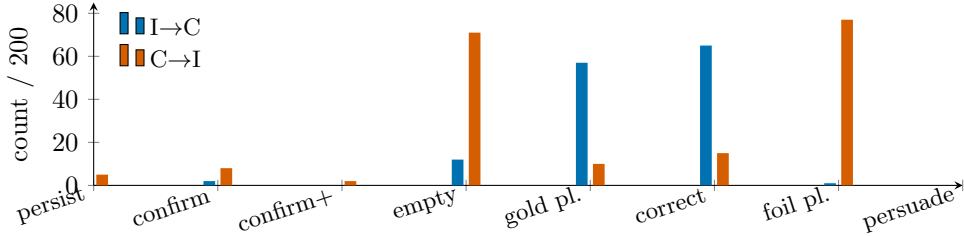

\centering
\figpayload
\caption{Speech act vs.\ payload on Qwen2.5-7B TriviaQA ($n=200$): I$\to$C and C$\to$I counts per T2 message. Gold and foil \emph{strings} move answers; re-asking and agreement do not. A bare foil statement flips more items than the templated \texttt{persuade} message. An empty challenge makes answers unstable without directing them}
\label{fig:payload}
\end{figure*}

Table~\ref{tab:g6} and Figure~\ref{fig:payload} are the controlled core of the study: T1 is identical and only the T2 message changes. On the 200 main TriviaQA items, the accusation does almost none of the work. Removing it leaves gold uptake nearly intact (\texttt{gold\_plain} converts 57 items against 65 for \texttt{correct}), and a bare foil pulls more answers away from gold than the templated \texttt{persuade} message (77 vs.\ 30). Removing the string instead removes the direction. \texttt{challenge\_empty} scatters answers both ways (71 C$\to$I, 12 I$\to$C), which looks like added randomness rather than an update. Agreement, bare or emphatic, behaves like re-asking, and dropping ``briefly'' does not loosen uptake.

The split does not depend on our exact T2 wording. Twenty-seven everyday paraphrases of the four T2 messages, applied to the same T1 histories, keep mean I$\to$C at $0.026$ when no string is named and $0.275$ when gold is named (foil C$\to$I $0.325$; Online Resource~1, Table~D1). Hedged, assistant-like phrasing does not replace the payload either. \emph{I think it's \{gold\}} still converts ($0.295$), while \emph{I think you might be off} does not ($0.045$). The same three ablations meet the same gates on NQ, HotpotQA, and WebQuestions. The pattern also holds on GSM8K under the short-answer protocol: \texttt{persist} converts $14/188$ T1-wrong items, and on the $174$ items still wrong, gold converts $0.437$ against $0.103$ for \emph{Try again}.

\begin{table}[t]
\centering
\caption{The payload split holds across models and datasets: an excerpt of the model$\times$dataset matrix (counts out of 200). The I$\to$C columns compare \texttt{persist} with \texttt{correct}; the C$\to$I columns compare \texttt{persist} with \texttt{persuade}. Stating gold converts $48$--$116$ items on every row under the contain matcher ($39$ for Mistral TriviaQA once answers that a model reading them judged not to endorse the gold are removed; Online Resource~1, Sect.~I); re-asking converts $0$--$11$. Qwen3 is omitted (truncated thinking cells). Llama is scored before the leaked role header (Section~\ref{sec:setup}). $\mathrm{P}_{\mathrm{err}}$ holds on all of these units}
\label{tab:factorial}
\footnotesize
\begin{tabular}{llrrrr}
\toprule
 & & \multicolumn{2}{c}{I$\to$C} & \multicolumn{2}{c}{C$\to$I} \\
\cmidrule(lr){3-4}\cmidrule(lr){5-6}
Model & Data & persist & correct & persist & persuade \\
\midrule
Qwen2.5-7B & TriviaQA & 0 & 65 & 5 & 30 \\
Qwen2.5-7B & NQ & 3 & 108 & 2 & 21 \\
Qwen2.5-7B & HotpotQA & 0 & 116 & 1 & 20 \\
Qwen2-7B & TriviaQA & 5 & 71 & 4 & 46 \\
Mistral-7B & TriviaQA & 2 & 49 & 0 & 23 \\
Llama-3-8B & TriviaQA & 8 & 48 & 1 & 56 \\
Phi-4 & TriviaQA & 0 & 50 & 0 & 8 \\
Qwen2.5-14B & TriviaQA & 2 & 51 & 1 & 22 \\
Qwen2.5-3B & TriviaQA & 6 & 84 & 12 & 27 \\
Llama-3-8B & NQ & 11 & 107 & 6 & 34 \\
Phi-4 & NQ & 1 & 109 & 2 & 10 \\
Mistral-7B & HotpotQA & 2 & 108 & 10 & 19 \\
\bottomrule
\end{tabular}
\end{table}

\paragraph{Generalization across the grid.}
In every retained full-size cell (all except PopQA, the 64-token Qwen3 cells, and the $n{\approx}20$ turn-depth subsamples), I$\to$C under \texttt{correct} is at least four times that under \texttt{persist} or \texttt{confirm} (Table~\ref{tab:factorial}). The closest cases are Llama-3-8B TriviaQA ($48$ vs.\ $11$) and GSM8K ($76$ vs.\ $18$ for \emph{Try again}). Across cells, I$\to$C under \texttt{persist} has a median of $2/200$ (minimum $0$). An earlier \texttt{persist} I$\to$C of $40$ came from the truncated Qwen3 cell that we withdraw in Online Resource~1, Sect.~C. Llama-3-8B misses $\mathrm{P}_{\mathrm{gold}}$ on TriviaQA and NQ because re-asking and agreement convert slightly more than the gates allow (\texttt{persist} I$\to$C $8$ and $11/200$; \texttt{confirm} $11$ and $10/200$). This is still far below gold's $48$ and $107$. Phi-4 barely moves under \texttt{persuade} (accuracy $0.70\to0.71$ on TriviaQA) but still adopts gold (I$\to$C $=50$, C$\to$I $=0$). Qwen2.5 and Llama-3-8B mostly copy the payload; Mistral-7B restates it inside a longer sentence, sometimes only to reject it. Copy rates and the model-read audit are in Online Resource~1, Sect.~I.

\section{Four measures, one transcript}\label{sec:inst}\label{sec:instruments}

\begin{table}[t]
\centering
\caption{Mean change from T1 to T2 ($\Delta=$ T2$-$T1) of each measure on Qwen2.5-7B TriviaQA ($n=200$). Values split by what the answer did are given in the text}
\label{tab:inst}
\small
\begin{tabular}{lrrrr}
\toprule
 & $\Delta$verb & $\Delta$ll & $\Delta$probe & $\Delta P(\mathrm{True})$ \\
\midrule
\texttt{persist} & $-0.10$ & $+0.14$ & $+0.15$ & $+0.007$ \\
\texttt{confirm} & $-0.10$ & $+0.14$ & $+0.15$ & $+0.002$ \\
\texttt{persuade} & $-0.17$ & $+0.02$ & $-0.03$ & $-0.258$ \\
\texttt{correct} & $-0.15$ & $+0.09$ & $+0.17$ & $-0.003$ \\
\bottomrule
\end{tabular}
\end{table}

\begin{figure}[t]
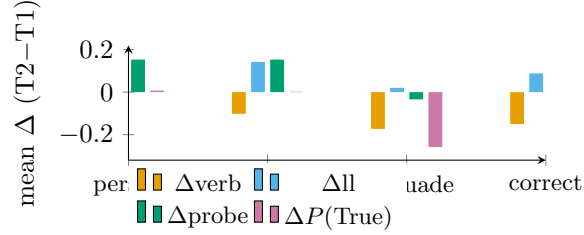

\centering
\figinstruments
\caption{Mean T2$-$T1 change of four measures on one Qwen2.5-7B TriviaQA set. Under \texttt{persist}, verbalized confidence falls, likelihood and the probe rise, and $P(\mathrm{True})$ stays still. Under \texttt{persuade}, $P(\mathrm{True})$ collapses while the probe on stay-wrong items still rises (Table~\ref{tab:inst})}
\label{fig:inst}
\end{figure}

The four measures disagree in sign in every row of Table~\ref{tab:inst} and Figure~\ref{fig:inst}. Under \texttt{correct}, verbalized confidence falls while accuracy climbs to $0.845$. The disagreement remains when we split items by what the answer did. Under \texttt{persist}, $P(\mathrm{True})$ barely moves on items that stayed wrong ($+0.012$), while the probe rises by $+0.38$ on the same items. Under \texttt{persuade}, $P(\mathrm{True})$ falls most where the answer flipped (C$\to$I $-0.845$, stay-wrong $-0.171$), yet the probe on stay-wrong items still rises ($+0.27$). Under \texttt{correct}, $P(\mathrm{True})$ moves both ways at once (I$\to$C $+0.148$, C$\to$I $-0.411$) around a mean of $\approx 0$, while the probe rises by $+0.64$ on I$\to$C items and $+0.27$ on the 16 stay-wrong items. No single ECE curve can capture that.

\paragraph{The drop in verbalized confidence is model-specific.}
The drop survives two obvious explanations. Residualizing on answer length leaves a mean $\Delta$ of $-0.13$, and the word ``briefly'' is not responsible, since \texttt{correct\_long} still converts. It does not survive a change of model. $\mathrm{P}_{\mathrm{verb}}$ holds on $6/6$ Qwen2.5-7B dataset units and on no other family once untrusted parses are dropped. Phi-4 moves the other way on all four core messages. Qwen2.5-3B and 14B fail on \texttt{confirm} ($+0.013$, $+0.044$). Llama-3's $-0.537$ is flagged \texttt{VERB\_UNTRUSTED} because its values pile up at $0$. Qwen3 fails to parse on 96\% of truncated rows because the \texttt{CONF} field sits inside \texttt{<think>}.

Under \texttt{persist} on Qwen2.5-7B TriviaQA, the most common sign pattern is likelihood rising while verbalized confidence does not rise ($172/200$ in the pilot cell; $51$ with a strict drop). In that cell, likelihood (the quantity MTCal calibrates) and verbalized confidence point in opposite directions on the same tokens.

\paragraph{Other models (post hoc).}
On Qwen2.5-14B, Mistral-7B-v0.3, Phi-4, and Qwen2-7B (not preregistered; Online Resource~1, Sect.~J), likelihood still rises under \texttt{persist} except on Mistral, $P(\mathrm{True})$ stays within $0.03$, and the all-item $\Delta$probe is negative on three of the four. The exact pattern in Table~\ref{tab:inst} is a Qwen2.5-7B result; what holds across models is the disagreement.

\paragraph{The same split appears in ECE.}
Table~\ref{tab:ece} scores all four measures against T1 and T2 correctness on one transcript. Under \texttt{persuade}, verbalized ECE \emph{falls} ($0.50\to0.38$), likelihood ECE nearly doubles ($0.23\to0.41$), and $P(\mathrm{True})$ ECE is flat ($0.22\to0.20$). The split in Figure~\ref{fig:ontology} therefore also appears in the field's standard metric. A study has to say which measure it calibrates.

The drop in verbalized ECE is not better calibration. Stated confidence stays about $0.3$ below accuracy at both turns ($0.28$ vs.\ $0.59$, then $0.10$ vs.\ $0.44$). AUROC (how well a score ranks correct above wrong answers) shows the same. At T1, verbalized confidence is close to uninformative ($0.564$), while likelihood, the probe, and $P(\mathrm{True})$ all sit near $0.85$. At T2 the order changes: verbalized confidence rises to $0.729$ and likelihood drops to $0.630$. Which measure a study happens to elicit decides how calibrated the result looks, and also whether it ranks correct answers at all. A post hoc rerun reproduces every cell of Tables~\ref{tab:inst} and~\ref{tab:ece} within $\pm0.002$. An older output file listed verbalized ECE as $0.582\to0.435$; it divided confidences that were already in $[0,1]$ by 100 again, and the values above are the corrected ones (Online Resource~1, Sect.~J).

The probe's own drop in AUROC ($0.848\to0.649$) is not noise. It is the persistence effect of Section~\ref{sec:commit} seen in calibration terms. A readout that rises on items that stay wrong (Table~\ref{tab:inst}) must lose discrimination against correctness once the user has spoken, because it now follows the answer the model keeps rather than what is true. In the same turn, the probe becomes a worse readout of truth and a better readout of answer persistence, and a single ``confidence'' number hides this.

\begin{table}[t]
\centering
\caption{ECE (10 bins) and AUROC against correctness on one transcript: Qwen2.5-7B TriviaQA, \texttt{persuade}, common $n{=}195$ (items whose verbalized confidence parses at both turns). Both probe columns are out-of-fold under GroupKFold over items: each fold's T1-fitted map scores that fold's own held-out T1 and T2 states, so the row is estimated the same way at both turns. The in-sample figures (ECE ${\le}0.002$, AUROC $1.000$) measure the regression on its own training set and are not calibration numbers. Verbalized ECE falls while mean confidence stays about $0.3$ below accuracy, so it is not a repair}
\label{tab:ece}
\small
\begin{tabular}{lcccc}
\toprule
& \multicolumn{2}{c}{ECE} & \multicolumn{2}{c}{AUROC} \\
\cmidrule(lr){2-3}\cmidrule(lr){4-5}
Measure & T1 & T2 & T1 & T2 \\
\midrule
Verbalized & $0.497$ & $0.375$ & $0.564$ & $0.729$ \\
Likelihood & $0.230$ & $0.411$ & $0.852$ & $0.630$ \\
Probe (held out) & $0.118$ & $0.260$ & $0.848$ & $0.649$ \\
$P(\mathrm{True})$ & $0.220$ & $0.203$ & $0.852$ & $0.862$ \\
\bottomrule
\end{tabular}
\end{table}

\section{Answer persistence, not truth}\label{sec:commit}

On Qwen2.5-7B, the frozen $v_{\mathrm{acc}}$ is a weak readout of truth and a consistent correlate of answer persistence. The post hoc controls below, including a rerun on held-out questions, limit the strong form of this result to Qwen2.5-7B, with Llama-3-8B on cut answers as support.

\paragraph{Direction and size.}
On all 24 independent units, stay-wrong $\Delta$probe is positive under both \texttt{persist} and \texttt{confirm} ($\mathrm{P}_{\mathrm{err}}=24/24$). These are in-sample numbers: the probe scores questions it was fitted on. On held-out questions the sign survives on Qwen2.5-7B and Llama-3-8B and fails on Qwen2.5-14B, Mistral, and Phi-4 (see \emph{Held-out questions} below). A sign count is the weakest form of this result, so we also report magnitudes. Across the 32 unit$\times$message cells with item-level states, the median of the mean $\Delta$probe is $0.264$ (interquartile range $[0.127, 0.384]$, range $[0.025, 0.758]$), and each cell's own bootstrap $95\%$ interval excludes zero ($32/32$). These cells are TriviaQA for seven models (Qwen3 in its truncated cell) and Qwen2.5-7B on five further slices. Six of the 32 are seed or configuration copies of the main Qwen2.5-7B cell. Sign counts over all $113$ cells are in Online Resource~1, Sect.~J. An MLP probe, a probe refitted on \texttt{persist} T2, and layer and token ablations keep the sign (Online Resource~1, Sect.~J).

So the probe rises when the model \emph{keeps a wrong answer} and when it \emph{adopts gold}, and it falls when a correct T1 answer is pulled away (\texttt{persuade} C$\to$I $\Delta$probe $-0.35$ in the unique-fit follow-up). It goes with the current answer, not with a latent true/false feature~\cite{marks2023geometry}.

\paragraph{Post hoc controls.}
Four checks were defined in a note written before the numbers and were not preregistered (Online Resource~1, Sect.~J). (a)~A probe fitted on shuffled T1 labels (100 seeds) stays near $0$ on Qwen2.5-7B, against $0.383$ for the real probe; on Qwen2.5-14B \texttt{persist}, Mistral, and Phi-4, $29$--$59\%$ of shuffles reach the real value. (b)~On Qwen2.5-7B the rise does not need a changed answer (identical answers $0.397$, $n=67$; changed $0.318$, $n=14$). (c)~Within stay-wrong items, $\Delta$probe and $\Delta$likelihood are only weakly related (Spearman $-0.09$ to $0.27$). (d)~A planned re-read without dialogue history is not interpretable: re-scoring the T1 answer did not reproduce the stored states (cosine $0.81$, below the required $0.95$, even after reusing the generated token ids), so we draw no conclusion from it (Online Resource~1, Sect.~J).

\paragraph{Held-out questions (post hoc).}
We refitted the probe in five folds grouped by question, so it never saw a test question at any turn, and scored stay-wrong items on the held-out folds only (Online Resource~1, Table~J10). On Qwen2.5-7B, $\Delta s$ is $0.321$ [$0.255$, $0.389$] under \texttt{persist} and $0.324$ [$0.252$, $0.395$] under \texttt{confirm} (in sample $0.383$, $0.384$), and no shuffled-label probe reaches it. Llama-3-8B, with states recomputed on cut answers, gives $0.201$ [$0.131$, $0.275$] and $0.119$ [$0.051$, $0.192$], also above every shuffle. On the other models the result does not hold. Qwen2-7B drops to $0.067$ [$-0.003$, $0.136$] and $0.019$ [$-0.058$, $0.090$]. Mistral and Phi-4 turn negative. Qwen2.5-14B is negative under \texttt{persist} ($-0.100$) and positive under \texttt{confirm} ($0.117$ [$0.019$, $0.212$]). The held-out probe still ranks T1 correctness on every model (AUROC $0.69$--$0.85$).

\subsection{Stay-wrong errors are hard to fix}\label{sec:l2}
Take the 81 Qwen2.5-7B TriviaQA items that are wrong at T1 and still wrong after \texttt{persist} at T2 (80 distinct questions; one appears twice in the slice). Further \texttt{persist} turns convert one of them by T3 (accuracy on stay-wrong items $=0.012$), two by T5 ($0.025$), and three by T10 ($0.037$). A reset at T3 (a fresh chat with only the original question) converts none. A T3 that continues the chat and offers abstention (\emph{If you are not confident \ldots\ reply \texttt{ABSTAIN}}) draws abstention on $94\%$ of stay-wrong and $62\%$ of stay-correct items. Every stay-wrong item that still answers is wrong, so coverage on stay-wrong items is $0.06$. Gating on the probe ($\mathrm{probe}_{T2}>\mathrm{probe}_{T1}$, true for 79 of the 80 distinct stay-wrong questions) does not change these accuracies. Re-asking, wiping the transcript, and offering abstention do not repair these errors.

A gold string at T3 does (Table~\ref{tab:t3sw}). Every gold phrasing recovers $0.75$--$0.95$ of the 81 items. Hedging does not lower it ($0.79$). A different gold alias recovers $0.77$ ($n_{\mathrm{SW}}=69$), with a copy rate of only $0.49$ for that surface form (main gold messages $0.61$--$0.71$), so the repair is not only echoing. The gold retrieved-note phrasing leaves stay-correct items intact (accuracy $0.965$, C$\to$I $4/114$).

\subsection{Preregistered replications}
Two preregistrations fixed models, items, messages, metrics, and a decision rule before any generation: a message without an answer string that reaches $\ge 0.15$ repair on stay-wrong items breaks the null for that model~\cite{prereg-ext,prereg-t3rep}. The first ran the turn-3 battery on Qwen's $n{=}500$ slice ($230$ stay-wrong items) and on the original Llama and Mistral items ($61$ and $58$; Online Resource~1, Sect.~E). The second regenerated Llama-3-8B and Mistral-7B-v0.3 on the same 500 items with correct stop tokens ($160$ and $139$ stay-wrong items) and added Qwen2.5-32B ($61$), all on GPU (Online Resource~1, Sect.~F).

Under the preregistered matcher, every observed rate without an answer string is below $0.15$ on every unit, so the decision rule never fires (Table~\ref{tab:t3sw}). That is weaker than excluding $0.15$, and several small cells did not (Wilson upper bounds up to $0.26$). A post hoc rerun on $298$ (Llama) and $269$ (32B) new stay-wrong items (Online Resource~1, Table~J11) now excludes it for re-asking, fresh chat, and \texttt{ABSTAIN} (at most $6/269$, upper bounds $\le0.048$). The empty elicit does not: $35/298$ [$0.086$, $0.159$] and $32/269$ [$0.086$, $0.163$]; pooled with earlier rows, Llama excludes $0.15$ ($50/458$, upper $0.141$) and 32B does not ($38/330$, upper $0.154$). Llama's highest arms in the first replication (up to $9/61=0.148$) come mostly from the leaked role header. In a post hoc check on the $54$ rows that are still stay-wrong after cutting answers at the header, fresh chat recovers $2/54$ ($0.04$) and the other arms without an answer string recover none (Online Resource~1, Table~E2). In the second replication the highest such arm is the empty elicit, at $6/61=0.098$ on 32B and $15/160=0.094$ on Llama (Wilson upper bounds $0.198$ and $0.149$; the 32B interval includes $0.15$). Mistral's gold rates ($0.72$/$0.71$) overstate uptake; read by a model, they are $0.55$/$0.25$.

One result goes against the null. With judge labels, and model-read labels for the rows where the matcher and judge disagree, Llama's empty elicit is $24/160=0.150$ [$0.103$, $0.213$], so the decision rule applied to those labels does \emph{not} replicate for Llama. But $9$ of the $24$ repairs were already correct at T1 and T2 under those labels; with stay-wrong defined by them (post hoc), the arm is $15/149=0.101$ [$0.062$, $0.159$] (Online Resource~1, Sect.~H).

Frozen $s_2$ does not predict which stay-wrong items gold leaves unrepaired (AUROC $0.39$, $0.49$, $0.62$ on Llama, Mistral, 32B; every interval contains $0.5$; Online Resource~1, Tables~E2 and~F3).

\subsection{Retrieval repairs only when it contains the answer}
Every gold arm above is an oracle: the user hands the model the right answer. The second preregistration therefore tested real evidence on the same stay-wrong items and on up to 150 stay-correct items per model. BM25~\cite{robertson2009bm25} ranks the sentences of each question's TriviaQA evidence documents, and T3 inserts one sentence into the retrieved-note message. The top-ranked sentence repairs $0.22$--$0.27$ of stay-wrong items, and how often it contains the answer explains that rate. Pooled over four models, $90/111$ items recover when it contains a gold alias and $20/330$ when it does not. Holding the ranking fixed, the best sentence that contains the answer repairs $0.67$--$0.76$, and the best sentence that lacks it repairs $0.02$--$0.12$. The difference is $0.56$--$0.73$, with every interval above $0$, and $0.55$--$0.61$ with model-read labels, under which Llama's no-answer sentence reaches $26/160=0.16$. Sentences without the answer are not neutral. They cut stay-correct accuracy from $1$ to $0.65$, $0.78$, $0.30$, and $0.78$.

\begin{table}[t]
\centering
\caption{T3 accuracy on \texttt{persist} stay-wrong items (contain matcher; rows 2--5 and the last column are preregistered). Re-ask: \texttt{persist} at T3 (main 81: $.025$ at T5); reset: fresh chat; abst.: \texttt{ABSTAIN} option; empty: elicit with no string; gold plain/retrieve are oracle arms; last column: best BM25 evidence sentence with/without a gold alias. $^{*}$Read by a model: .55/.25 and .63/.06. Intervals and other scorers are in Online Resource~1, Sects.~E--H}
\label{tab:t3sw}
\footnotesize
\setlength{\tabcolsep}{3pt}
\begin{tabular}{lrcccc}
\toprule
Stay-wrong unit & $n$ & re-ask/reset/abst. & empty & gold plain/retr. & answer/no answer \\
\midrule
Qwen2.5-7B main & 81 & .01/.00/.00 & .06 & .78/.93 & .67/.05 \\
Qwen2.5-7B $n{=}500$ & 230 & .03/.05/.00 & .03 & .92/.94 & -- \\
Llama-3-8B & 160 & .01/.00/.01 & .09 & .89/.91 & .68/.12 \\
Mistral-7B & 139 & .01/.00/.00 & .01 & .72/.71$^{*}$ & .76/.02$^{*}$ \\
Qwen2.5-32B & 61 & .00/.00/.00 & .10 & .92/.90 & .69/.11 \\
\bottomrule
\end{tabular}
\end{table}

\section{Implications for conversational systems}\label{sec:discussion}

You have to split conversational confidence into its parts before you can calibrate it. ECE@T on likelihood~\cite{zhang-etal-2026-mtcal} scores fluency, and $P(\mathrm{Sufficient})$~\cite{zhang-etal-2026-confidence-estimation} scores evidence. Neither measures uptake of the user's string or the persistence readout $s_t$. On one transcript, I$\to$C and C$\to$I counts show whether a user string overwrote the answer, $s_t$ (on Qwen2.5-7B) follows the answer the model keeps, $P(\mathrm{True})$ shows whether a contradiction landed, and likelihood shows fluency. Table~\ref{tab:change} lists what this means in practice.

\begin{table}[t]
\centering
\caption{Suggested changes to evaluation and system design if conversational confidence is not one latent belief}
\label{tab:change}
\small
\begin{tabular}{p{0.38\textwidth}p{0.54\textwidth}}
\toprule
Current practice & Suggested change \\
\midrule
Report ECE@T, verbalized-confidence AUROC, or $P(\mathrm{True})$ as \emph{the} multi-turn confidence number &
State which measure in Figure~\ref{fig:ontology} is scored. Scores on one transcript are not substitutes, and fixing one does not fix the others. \\
Bundle ``you are wrong'' with a gold or foil string &
Separate the speech act from the answer string; otherwise one cannot say what moved. \\
Reset the chat, or gate on a drop in verbalized confidence, to undo errors &
These act on the history and the stated confidence, not on $s_t$: accuracy on the 81 stay-wrong items stays $0$. A gold string at T3 repairs them; a foil string overwrites stay-correct answers (Section~\ref{sec:l2}). \\
Add retrieval and expect repair &
Heuristic, not validated end to end: check that the evidence contains the answer string. A retrieved BM25 sentence that contains it repairs $0.67$--$0.76$ of stay-wrong items, one without it $0.02$--$0.12$ (four models). \\
\bottomrule
\end{tabular}
\end{table}

\paragraph{For evaluation.}
Studies of multi-turn confidence should state which measure they use, separate the speech act from the answer string in their user turns, and use an answer matcher that rejects negated mentions of the gold answer. A calibration result for one measure is not evidence about another.

\paragraph{For system design.}
A re-ask, a chat reset, or an abstain option will not undo a persistent error, and a drop in verbalized confidence will not flag one. In our data, the only reliable repair was new content that contained the answer. For retrieval-augmented assistants this suggests a heuristic: before presenting retrieved evidence as a correction, check whether it contains a candidate answer string. In our data, evidence without one rarely repairs an error and can break up to $0.70$ of answers that were correct. In a post hoc end-to-end test (BM25 passages, TriviaQA and HotpotQA, three models; Online Resource~1, Table~J12), the gold-based gate cut harm on TriviaQA while keeping most repairs (Llama, $k=1$: harm $0.179\to0.038$, net per 100 items $15.9\to27.4$); on HotpotQA the gain was mixed. A gate that checks the model's own answer instead of the gold repaired only $0.02$--$0.12$ of errors, and dense retrieval and a judge check were not run. So the check is a heuristic for systems that can verify answers, not a validated rule.

\section{Limitations}\label{sec:limits}

\paragraph{What we do not claim.}
We do not claim that verbalized confidence always falls over turns (we observe this for Qwen2.5-7B only), that hidden states contain a truth direction, or that we outperform MTCal~\cite{zhang-etal-2026-mtcal} or $P(\mathrm{Sufficient})$~\cite{zhang-etal-2026-confidence-estimation} on ECE. We do not claim to be first to show that verbalized confidence and correctness-tracking signals come apart, or that confidence can track commitment; single-turn studies showed both~\cite{kumaran2026commitment,miao2026confidence}. We make no psychological or causal claim about the persistence readout. We propose no new uncertainty estimator.

\paragraph{Task scope.}
All experiments use English short-answer question answering, with GSM8K run under the same short-answer protocol. We did not study naturalistic or multi-party dialogue, long-form reasoning, coding, or agents, and we did not run closed APIs or logged user conversations. Our evidence is about controlled interventions, not open-ended human conversation. The turn-3 battery contains no reasoning prompt, so we do not claim that step-by-step reconsideration cannot repair persistent errors. The factorial design is not a formal causal-identification argument.

\paragraph{Model scale and families.}
The main grid covers 3B--14B open models. Above 14B we ran only Qwen2.5-32B, and only the stay-wrong battery, its probe, and retrieval. The drop in verbalized confidence is a Qwen2.5-7B result in our grid, and we do not generalize it to other families. Qwen3 needs the answer extracted after \texttt{</think>} (Online Resource~1, Sect.~C).

\paragraph{Sample size of the key cell.}
The central stay-wrong analysis on Qwen2.5-7B rests on $n=81$ items (80 distinct questions) from one dataset. The preregistered replications extend this to $230$, $160$, $139$, and $61$ items on four models, but the turn-3 battery and the retrieval study cover TriviaQA only. 

\paragraph{The persistence readout is defined by a probe.}
$s_t$ is an operational quantity: a frozen readout fitted on T1 states. We make claims about how it behaves when we intervene, not about what the model ``believes''. We report no activation-steering or patching results, so we cannot say whether $s_t$ is causally involved in producing the answer. The single-turn studies that do steer~\cite{kumaran2026commitment,miao2026confidence} use different readouts and tasks, so their causal results do not transfer to ours. Llama-3-8B probe states in the main analyses were pooled over the raw generated answer tokens, including any repetition after the leaked role header. In a post hoc rerun on cut answers, stay-wrong $\Delta$probe stays positive on every Llama unit we could test but is much smaller (TriviaQA \texttt{persist} $0.865\to0.333$, NQ $0.399\to0.034$, HotpotQA $0.238\to0.030$; Online Resource~1, Sect.~J). Llama magnitudes from the main analyses should be read as upper bounds. The post hoc shuffled-label control (Section~\ref{sec:commit}) and the held-out-question rerun (Section~\ref{sec:commit}) leave the stay-wrong rise standing on Qwen2.5-7B, with Llama-3-8B on cut answers as support. On Qwen2.5-14B, Mistral, Phi-4, and Qwen2-7B it does not hold on held-out questions, so the in-sample $\mathrm{P}_{\mathrm{err}}=24/24$ overstates how general it is.

\paragraph{Answer matching is fragile.}
The primary contain matcher credits answers that name the gold only to reject it, which inflates Mistral's gold uptake in particular. PopQA is omitted because the matcher and output format failed. Answer labels come from the matcher, an LLM judge, and model reads; post hoc, two people labelled a blind $300$-row V2 sample with Cohen's $\kappa{=}0.86$ on $272$ clear pairs (Online Resource~1, Sect.~H). Replication and retrieval results are also reported under contain, judge, and model-read labels, and the Mistral results after a rejection audit (Online Resource~1, Sects.~H and~I). Under these alternative labels Llama's empty-elicit arm reaches the $0.15$ threshold (Section~\ref{sec:l2}). Table~\ref{tab:ece} reports ECE for all four measures on one message type of one model$\times$dataset cell. We do not report ECE as a grid-wide result, and we claim neither a win nor a loss against MTCal.

\section{Conclusion}\label{sec:concl}

In short-answer question answering, an LLM's confidence in a conversation is not one belief waiting for a better estimator. What the user supplies (an answer string) moves the answer. How the user phrases the turn (re-asking, agreeing, or accusing) does not. Four common confidence measures respond to different parts of the same turn, and on Qwen2.5-7B a frozen hidden-state readout follows the answer the model keeps rather than its truth. Errors that survive one re-ask are not undone by re-asking, resetting, or offering abstention. A correct answer string repairs them, including one inside real retrieved evidence. For practitioners our data suggest a heuristic that still needs end-to-end testing: before relying on evidence to correct a model, check that it contains the answer.

\backmatter

\bmhead{Supplementary information}
Online Resource 1 (ESM\_1.pdf) contains the supplementary appendices, Sects.~A--I (prompt wording, additional and robustness controls, the Qwen3 analysis, both preregistered replications, retrieval, scoring, and the rejection audit). Online Resource 2 (ESM\_2.zip) contains the three preregistration documents (PREREG.md, PREREG\_T3REP.md, PREREG\_TMLR.md) referred to in Section~\ref{sec:prereg}.


\section*{Statements and Declarations}

\begin{itemize}
\item \textbf{Funding:} No funding was received for conducting this study.
\item \textbf{Competing interests:} The authors have no competing interests to declare that are relevant to the content of this article.
\item \textbf{Ethics approval and consent to participate:} Not applicable. The study involves no human participants, human data, or animals; all questions and answers come from public question-answering datasets.
\item \textbf{Consent for publication:} Not applicable.
\item \textbf{Data availability:} The underlying QA datasets analysed in this study are publicly available: TriviaQA, Natural Questions (NQ-open), HotpotQA, TruthfulQA, WebQuestions, and GSM8K (see the references in Section~\ref{sec:setup}). The generated dialogues, model answers, labels, and analysis outputs that support the findings of this study are available from the corresponding author on reasonable request (\texttt{vinh.td225538@sis.hust.edu.vn}).
\item \textbf{Materials availability:} Not applicable.
\item \textbf{Code availability:} The code used to run the experiments and produce the tables and figures, together with the preregistration documents and their timestamped freezing commits, is available from the corresponding author on reasonable request (\texttt{vinh.td225538@sis.hust.edu.vn}).
\item \textbf{Author contributions:} Duc-Vinh Tran: conceptualization, methodology, software, investigation, formal analysis, writing (original draft and review and editing).
\end{itemize}

\bibliography{custom}


\begin{thebibliography}{44}
\ifx \bisbn   \undefined \def \bisbn  #1{ISBN #1}\fi
\ifx \binits  \undefined \def \binits#1{#1}\fi
\ifx \bauthor  \undefined \def \bauthor#1{#1}\fi
\ifx \batitle  \undefined \def \batitle#1{#1}\fi
\ifx \bjtitle  \undefined \def \bjtitle#1{#1}\fi
\ifx \bvolume  \undefined \def \bvolume#1{\textbf{#1}}\fi
\ifx \byear  \undefined \def \byear#1{#1}\fi
\ifx \bissue  \undefined \def \bissue#1{#1}\fi
\ifx \bfpage  \undefined \def \bfpage#1{#1}\fi
\ifx \blpage  \undefined \def \blpage #1{#1}\fi
\ifx \burl  \undefined \def \burl#1{\textsf{#1}}\fi
\ifx \doiurl  \undefined \def \doiurl#1{\url{https://doi.org/#1}}\fi
\ifx \betal  \undefined \def \betal{\textit{et al.}}\fi
\ifx \binstitute  \undefined \def \binstitute#1{#1}\fi
\ifx \binstitutionaled  \undefined \def \binstitutionaled#1{#1}\fi
\ifx \bctitle  \undefined \def \bctitle#1{#1}\fi
\ifx \beditor  \undefined \def \beditor#1{#1}\fi
\ifx \bpublisher  \undefined \def \bpublisher#1{#1}\fi
\ifx \bbtitle  \undefined \def \bbtitle#1{#1}\fi
\ifx \bedition  \undefined \def \bedition#1{#1}\fi
\ifx \bseriesno  \undefined \def \bseriesno#1{#1}\fi
\ifx \blocation  \undefined \def \blocation#1{#1}\fi
\ifx \bsertitle  \undefined \def \bsertitle#1{#1}\fi
\ifx \bsnm \undefined \def \bsnm#1{#1}\fi
\ifx \bsuffix \undefined \def \bsuffix#1{#1}\fi
\ifx \bparticle \undefined \def \bparticle#1{#1}\fi
\ifx \barticle \undefined \def \barticle#1{#1}\fi
\bibcommenthead
\ifx \bconfdate \undefined \def \bconfdate #1{#1}\fi
\ifx \botherref \undefined \def \botherref #1{#1}\fi
\ifx \url \undefined \def \url#1{\textsf{#1}}\fi
\ifx \bchapter \undefined \def \bchapter#1{#1}\fi
\ifx \bbook \undefined \def \bbook#1{#1}\fi
\ifx \bcomment \undefined \def \bcomment#1{#1}\fi
\ifx \oauthor \undefined \def \oauthor#1{#1}\fi
\ifx \citeauthoryear \undefined \def \citeauthoryear#1{#1}\fi
\ifx \endbibitem  \undefined \def \endbibitem {}\fi
\ifx \bconflocation  \undefined \def \bconflocation#1{#1}\fi
\ifx \arxivurl  \undefined \def \arxivurl#1{\textsf{#1}}\fi
\csname PreBibitemsHook\endcsname

\bibitem[\protect\citeauthoryear{Zhang
  et~al.}{2026a}]{zhang-etal-2026-confidence-estimation}
\begin{bchapter}
\bauthor{\bsnm{Zhang}, \binits{C.}},
\bauthor{\bsnm{Yang}, \binits{R.}},
\bauthor{\bsnm{Zhu}, \binits{X.}},
\bauthor{\bsnm{Li}, \binits{C.}},
\bauthor{\bsnm{Hu}, \binits{T.}},
\bauthor{\bsnm{Dong}, \binits{Y.R.}},
\bauthor{\bsnm{Yang}, \binits{D.}},
\bauthor{\bsnm{Collier}, \binits{N.}}:
\bctitle{Confidence estimation for {LLM}s in multi-turn interactions}.
In: \bbtitle{Findings of the Association for Computational Linguistics: ACL
  2026},
pp. \bfpage{25661}--\blpage{25676}.
\bpublisher{Association for Computational Linguistics},
\blocation{San Diego, California, United States}
(\byear{2026}).
\doiurl{10.18653/v1/2026.findings-acl.1280} .
\burl{https://aclanthology.org/2026.findings-acl.1280/}
\end{bchapter}
\endbibitem

\bibitem[\protect\citeauthoryear{Zhang et~al.}{2026b}]{zhang-etal-2026-mtcal}
\begin{bchapter}
\bauthor{\bsnm{Zhang}, \binits{Z.}},
\bauthor{\bsnm{Li}, \binits{C.}},
\bauthor{\bsnm{Liu}, \binits{X.}},
\bauthor{\bsnm{Shen}, \binits{C.}},
\bauthor{\bsnm{Liu}, \binits{Z.}},
\bauthor{\bsnm{Patras}, \binits{I.}}:
\bctitle{Confidence should be calibrated more than one turn deep}.
In: \bbtitle{Proceedings of the 64th Annual Meeting of the Association for
  Computational Linguistics (Volume 1: Long Papers)},
pp. \bfpage{38578}--\blpage{38594}.
\bpublisher{Association for Computational Linguistics},
\blocation{San Diego, California, United States}
(\byear{2026}).
\doiurl{10.18653/v1/2026.acl-long.1787} .
\burl{https://aclanthology.org/2026.acl-long.1787/}
\end{bchapter}
\endbibitem

\bibitem[\protect\citeauthoryear{Kumaran}{2026}]{kumaran2026commitment}
\begin{botherref}
\oauthor{\bsnm{Kumaran}, \binits{D.}}:
Reported Confidence in {LLMs} Tracks Commitment More Than Correctness.
arXiv preprint arXiv:2606.29490
(2026).
\url{https://arxiv.org/abs/2606.29490}
\end{botherref}
\endbibitem

\bibitem[\protect\citeauthoryear{Kadavath et~al.}{2022}]{kadavath2022mostly}
\begin{botherref}
\oauthor{\bsnm{Kadavath}, \binits{S.}},
\oauthor{\bsnm{Conerly}, \binits{T.}},
\oauthor{\bsnm{Askell}, \binits{A.}},
\oauthor{\bsnm{Henighan}, \binits{T.}},
\oauthor{\bsnm{Drain}, \binits{D.}},
\oauthor{\bsnm{Perez}, \binits{E.}},
\oauthor{\bsnm{Schiefer}, \binits{N.}},
\oauthor{\bsnm{Hatfield-Dodds}, \binits{Z.}},
\oauthor{\bsnm{DasSarma}, \binits{N.}},
\oauthor{\bsnm{Mann}, \binits{E.}}, et al.:
Language models (mostly) know what they know.
arXiv preprint arXiv:2207.05221
(2022)
\end{botherref}
\endbibitem

\bibitem[\protect\citeauthoryear{Miao and Ungar}{2026}]{miao2026confidence}
\begin{botherref}
\oauthor{\bsnm{Miao}, \binits{M.M.}},
\oauthor{\bsnm{Ungar}, \binits{L.}}:
Closing the Confidence-Faithfulness Gap in Large Language Models.
arXiv preprint arXiv:2603.25052
(2026).
\url{https://arxiv.org/abs/2603.25052}
\end{botherref}
\endbibitem

\bibitem[\protect\citeauthoryear{Perez et~al.}{2023}]{perez2022sycophancy}
\begin{bchapter}
\bauthor{\bsnm{Perez}, \binits{E.}},
\bauthor{\bsnm{Ringer}, \binits{S.}},
\bauthor{\bsnm{Luko{\v{s}}i{\=u}t{\.e}}, \binits{K.}},
\bauthor{\bsnm{Nguyen}, \binits{K.}},
\bauthor{\bsnm{Chen}, \binits{E.}},
\bauthor{\bsnm{Heiner}, \binits{S.}},
\bauthor{\bsnm{Pettit}, \binits{C.}},
\bauthor{\bsnm{Olsson}, \binits{C.}},
\bauthor{\bsnm{Kundu}, \binits{S.}},
\bauthor{\bsnm{Kadavath}, \binits{S.}}, \betal:
\bctitle{Discovering language model behaviors with model-written evaluations}.
In: \bbtitle{Findings of the Association for Computational Linguistics: ACL
  2023}
(\byear{2023})
\end{bchapter}
\endbibitem

\bibitem[\protect\citeauthoryear{Sharma et~al.}{2024}]{sharma2023sycophancy}
\begin{bchapter}
\bauthor{\bsnm{Sharma}, \binits{M.}},
\bauthor{\bsnm{Tong}, \binits{M.}},
\bauthor{\bsnm{Korbak}, \binits{T.}},
\bauthor{\bsnm{Duvenaud}, \binits{D.}},
\bauthor{\bsnm{Askell}, \binits{A.}},
\bauthor{\bsnm{Bowman}, \binits{S.R.}},
\bauthor{\bsnm{Cheng}, \binits{N.}},
\bauthor{\bsnm{Durmus}, \binits{E.}},
\bauthor{\bsnm{Hatfield-Dodds}, \binits{Z.}},
\bauthor{\bsnm{Johnston}, \binits{S.R.}}, \betal:
\bctitle{Towards understanding sycophancy in language models}.
In: \bbtitle{International Conference on Learning Representations (ICLR)}
(\byear{2024})
\end{bchapter}
\endbibitem

\bibitem[\protect\citeauthoryear{Wei et~al.}{2023}]{wei2023simple}
\begin{botherref}
\oauthor{\bsnm{Wei}, \binits{J.}},
\oauthor{\bsnm{Huang}, \binits{D.}},
\oauthor{\bsnm{Lu}, \binits{Y.}},
\oauthor{\bsnm{Zhou}, \binits{D.}},
\oauthor{\bsnm{Le}, \binits{Q.V.}}:
Simple synthetic data reduces sycophancy in large language models.
arXiv preprint arXiv:2308.03958
(2023)
\end{botherref}
\endbibitem

\bibitem[\protect\citeauthoryear{Laban et~al.}{2023}]{laban2023flipflop}
\begin{botherref}
\oauthor{\bsnm{Laban}, \binits{P.}},
\oauthor{\bsnm{Murakhovs'ka}, \binits{L.}},
\oauthor{\bsnm{Xiong}, \binits{C.}},
\oauthor{\bsnm{Wu}, \binits{C.-S.}}:
Are you sure? challenging {LLMs} leads to performance drops in the {FlipFlop}
  experiment.
arXiv preprint arXiv:2311.08596
(2023)
\end{botherref}
\endbibitem

\bibitem[\protect\citeauthoryear{Xie et~al.}{2024}]{xie2024askagain}
\begin{bchapter}
\bauthor{\bsnm{Xie}, \binits{Q.}},
\bauthor{\bsnm{Wang}, \binits{Z.}},
\bauthor{\bsnm{Feng}, \binits{Y.}},
\bauthor{\bsnm{Xia}, \binits{R.}}:
\bctitle{Ask again, then fail: Large language models' vacillations in
  judgment}.
In: \bbtitle{Proceedings of the 62nd Annual Meeting of the Association for
  Computational Linguistics (Volume 1: Long Papers)}.
\bpublisher{Association for Computational Linguistics},
\blocation{Bangkok, Thailand}
(\byear{2024}).
\doiurl{10.18653/v1/2024.acl-long.577}
\end{bchapter}
\endbibitem

\bibitem[\protect\citeauthoryear{Wang et~al.}{2023}]{wang2023defend}
\begin{bchapter}
\bauthor{\bsnm{Wang}, \binits{B.}},
\bauthor{\bsnm{Yue}, \binits{X.}},
\bauthor{\bsnm{Sun}, \binits{H.}}:
\bctitle{Can {ChatGPT} defend its belief in truth? evaluating {LLM} reasoning
  via debate}.
In: \bbtitle{Findings of the Association for Computational Linguistics: EMNLP
  2023}
(\byear{2023})
\end{bchapter}
\endbibitem

\bibitem[\protect\citeauthoryear{Huang et~al.}{2024}]{huang2024selfcorrect}
\begin{bchapter}
\bauthor{\bsnm{Huang}, \binits{J.}},
\bauthor{\bsnm{Chen}, \binits{X.}},
\bauthor{\bsnm{Mishra}, \binits{S.}},
\bauthor{\bsnm{Zheng}, \binits{H.S.}},
\bauthor{\bsnm{Yu}, \binits{A.W.}},
\bauthor{\bsnm{Song}, \binits{X.}},
\bauthor{\bsnm{Zhou}, \binits{D.}}:
\bctitle{Large language models cannot self-correct reasoning yet}.
In: \bbtitle{International Conference on Learning Representations (ICLR)}
(\byear{2024})
\end{bchapter}
\endbibitem

\bibitem[\protect\citeauthoryear{Laban et~al.}{2026}]{laban2025lost}
\begin{bchapter}
\bauthor{\bsnm{Laban}, \binits{P.}},
\bauthor{\bsnm{Hayashi}, \binits{H.}},
\bauthor{\bsnm{Zhou}, \binits{Y.}},
\bauthor{\bsnm{Neville}, \binits{J.}}:
\bctitle{{LLMs} get lost in multi-turn conversation}.
In: \bbtitle{International Conference on Learning Representations (ICLR)}
(\byear{2026}).
\burl{https://arxiv.org/abs/2505.06120}
\end{bchapter}
\endbibitem

\bibitem[\protect\citeauthoryear{Lin et~al.}{2022}]{lin2022teaching}
\begin{botherref}
\oauthor{\bsnm{Lin}, \binits{S.}},
\oauthor{\bsnm{Hilton}, \binits{J.}},
\oauthor{\bsnm{Evans}, \binits{O.}}:
Teaching models to express their uncertainty in words.
Transactions on Machine Learning Research
(2022)
\end{botherref}
\endbibitem

\bibitem[\protect\citeauthoryear{Xiong et~al.}{2024}]{xiong2024verbalized}
\begin{bchapter}
\bauthor{\bsnm{Xiong}, \binits{M.}},
\bauthor{\bsnm{Hu}, \binits{Z.}},
\bauthor{\bsnm{Lu}, \binits{X.}},
\bauthor{\bsnm{Li}, \binits{Y.}},
\bauthor{\bsnm{Fu}, \binits{J.}},
\bauthor{\bsnm{He}, \binits{J.}},
\bauthor{\bsnm{Hooi}, \binits{B.}}:
\bctitle{Can {LLM}s express their uncertainty? an empirical evaluation of
  confidence elicitation in {LLM}s}.
In: \bbtitle{International Conference on Learning Representations (ICLR)}
(\byear{2024})
\end{bchapter}
\endbibitem

\bibitem[\protect\citeauthoryear{Tian et~al.}{2023}]{tian2023just}
\begin{bchapter}
\bauthor{\bsnm{Tian}, \binits{K.}},
\bauthor{\bsnm{Mitchell}, \binits{E.}},
\bauthor{\bsnm{Zhou}, \binits{A.}},
\bauthor{\bsnm{Sharma}, \binits{A.}},
\bauthor{\bsnm{Rafailov}, \binits{R.}},
\bauthor{\bsnm{Yao}, \binits{H.}},
\bauthor{\bsnm{Finn}, \binits{C.}},
\bauthor{\bsnm{Manning}, \binits{C.D.}}:
\bctitle{Just ask for calibration: Strategies for eliciting calibrated
  confidence scores from language models fine-tuned with human feedback}.
In: \bbtitle{Proceedings of the 2023 Conference on Empirical Methods in Natural
  Language Processing (EMNLP)}
(\byear{2023})
\end{bchapter}
\endbibitem

\bibitem[\protect\citeauthoryear{Geng et~al.}{2024}]{geng2024survey}
\begin{bchapter}
\bauthor{\bsnm{Geng}, \binits{J.}},
\bauthor{\bsnm{Cai}, \binits{F.}},
\bauthor{\bsnm{Wang}, \binits{Y.}},
\bauthor{\bsnm{Koeppl}, \binits{H.}},
\bauthor{\bsnm{Nakov}, \binits{P.}},
\bauthor{\bsnm{Gurevych}, \binits{I.}}:
\bctitle{A survey of confidence estimation and calibration in large language
  models}.
In: \bbtitle{Proceedings of the 2024 Conference of the North American Chapter
  of the Association for Computational Linguistics (NAACL)}
(\byear{2024})
\end{bchapter}
\endbibitem

\bibitem[\protect\citeauthoryear{Guo et~al.}{2017}]{guo2017calibration}
\begin{bchapter}
\bauthor{\bsnm{Guo}, \binits{C.}},
\bauthor{\bsnm{Pleiss}, \binits{G.}},
\bauthor{\bsnm{Sun}, \binits{Y.}},
\bauthor{\bsnm{Weinberger}, \binits{K.Q.}}:
\bctitle{On calibration of modern neural networks}.
In: \bbtitle{Proceedings of the 34th International Conference on Machine
  Learning}.
\bsertitle{Proceedings of Machine Learning Research},
vol. \bseriesno{70},
pp. \bfpage{1321}--\blpage{1330}.
\bpublisher{PMLR},
\blocation{Sydney, Australia}
(\byear{2017})
\end{bchapter}
\endbibitem

\bibitem[\protect\citeauthoryear{Azaria and
  Mitchell}{2023}]{azaria2023internal}
\begin{bchapter}
\bauthor{\bsnm{Azaria}, \binits{A.}},
\bauthor{\bsnm{Mitchell}, \binits{T.}}:
\bctitle{The internal state of an {LLM} knows when it's lying}.
In: \bbtitle{Findings of the Association for Computational Linguistics: EMNLP
  2023}
(\byear{2023})
\end{bchapter}
\endbibitem

\bibitem[\protect\citeauthoryear{Marks and Tegmark}{2024}]{marks2023geometry}
\begin{bchapter}
\bauthor{\bsnm{Marks}, \binits{S.}},
\bauthor{\bsnm{Tegmark}, \binits{M.}}:
\bctitle{The geometry of truth: Emergent linear structure in large language
  model representations of true/false datasets}.
In: \bbtitle{Conference on Language Modeling (COLM)}
(\byear{2024})
\end{bchapter}
\endbibitem

\bibitem[\protect\citeauthoryear{Burns et~al.}{2023}]{burns2023ccs}
\begin{bchapter}
\bauthor{\bsnm{Burns}, \binits{C.}},
\bauthor{\bsnm{Ye}, \binits{H.}},
\bauthor{\bsnm{Klein}, \binits{D.}},
\bauthor{\bsnm{Steinhardt}, \binits{J.}}:
\bctitle{Discovering latent knowledge in language models without supervision}.
In: \bbtitle{International Conference on Learning Representations (ICLR)}
(\byear{2023})
\end{bchapter}
\endbibitem

\bibitem[\protect\citeauthoryear{Orgad et~al.}{2025}]{orgad2025llmsknow}
\begin{bchapter}
\bauthor{\bsnm{Orgad}, \binits{H.}},
\bauthor{\bsnm{Toker}, \binits{M.}},
\bauthor{\bsnm{Gekhman}, \binits{Z.}},
\bauthor{\bsnm{Reichart}, \binits{R.}},
\bauthor{\bsnm{Szpektor}, \binits{I.}},
\bauthor{\bsnm{Kotek}, \binits{H.}},
\bauthor{\bsnm{Belinkov}, \binits{Y.}}:
\bctitle{{LLMs} know more than they show: On the intrinsic representation of
  {LLM} hallucinations}.
In: \bbtitle{International Conference on Learning Representations (ICLR)}
(\byear{2025})
\end{bchapter}
\endbibitem

\bibitem[\protect\citeauthoryear{Hewitt and Liang}{2019}]{hewitt2019control}
\begin{bchapter}
\bauthor{\bsnm{Hewitt}, \binits{J.}},
\bauthor{\bsnm{Liang}, \binits{P.}}:
\bctitle{Designing and interpreting probes with control tasks}.
In: \bbtitle{Proceedings of the 2019 Conference on Empirical Methods in Natural
  Language Processing (EMNLP)}
(\byear{2019})
\end{bchapter}
\endbibitem

\bibitem[\protect\citeauthoryear{Sarkar et~al.}{2026}]{sarkar2026contextual}
\begin{botherref}
\oauthor{\bsnm{Sarkar}, \binits{R.}},
\oauthor{\bsnm{Ramu}, \binits{P.}},
\oauthor{\bsnm{Rudinger}, \binits{R.}}:
Language Models Encode the Contextual Truth of Propositions.
arXiv preprint arXiv:2608.03035
(2026).
\url{https://arxiv.org/abs/2608.03035}
\end{botherref}
\endbibitem

\bibitem[\protect\citeauthoryear{Longpre et~al.}{2021}]{longpre2021entity}
\begin{bchapter}
\bauthor{\bsnm{Longpre}, \binits{S.}},
\bauthor{\bsnm{Perisetla}, \binits{K.}},
\bauthor{\bsnm{Chen}, \binits{A.}},
\bauthor{\bsnm{Ramesh}, \binits{N.}},
\bauthor{\bsnm{DuBois}, \binits{C.}},
\bauthor{\bsnm{Singh}, \binits{S.}}:
\bctitle{Entity-based knowledge conflicts in question answering}.
In: \bbtitle{Proceedings of the 2021 Conference on Empirical Methods in Natural
  Language Processing (EMNLP)}
(\byear{2021})
\end{bchapter}
\endbibitem

\bibitem[\protect\citeauthoryear{Xie et~al.}{2024}]{xie2024chameleon}
\begin{bchapter}
\bauthor{\bsnm{Xie}, \binits{J.}},
\bauthor{\bsnm{Zhang}, \binits{K.}},
\bauthor{\bsnm{Chen}, \binits{J.}},
\bauthor{\bsnm{Lou}, \binits{R.}},
\bauthor{\bsnm{Su}, \binits{Y.}}:
\bctitle{Adaptive chameleon or stubborn sloth: Revealing the behavior of large
  language models in knowledge conflicts}.
In: \bbtitle{International Conference on Learning Representations (ICLR)}
(\byear{2024})
\end{bchapter}
\endbibitem

\bibitem[\protect\citeauthoryear{Joren et~al.}{2025}]{joren2025sufficient}
\begin{bchapter}
\bauthor{\bsnm{Joren}, \binits{H.}},
\bauthor{\bsnm{Zhang}, \binits{J.}},
\bauthor{\bsnm{Ferng}, \binits{C.-S.}},
\bauthor{\bsnm{Juan}, \binits{D.-C.}},
\bauthor{\bsnm{Taly}, \binits{A.}},
\bauthor{\bsnm{Rashtchian}, \binits{C.}}:
\bctitle{Sufficient context: A new lens on retrieval augmented generation
  systems}.
In: \bbtitle{International Conference on Learning Representations (ICLR)}
(\byear{2025})
\end{bchapter}
\endbibitem

\bibitem[\protect\citeauthoryear{Lathkar}{2026}]{lathkar2026anchored}
\begin{botherref}
\oauthor{\bsnm{Lathkar}, \binits{A.B.}}:
Anchored Confabulation: Partial Evidence Non-Monotonically Amplifies Confident
  Hallucination in {LLMs}.
arXiv preprint arXiv:2604.25931
(2026).
\url{https://arxiv.org/abs/2604.25931}
\end{botherref}
\endbibitem

\bibitem[\protect\citeauthoryear{Kuhn et~al.}{2023}]{kuhn2023semantic}
\begin{bchapter}
\bauthor{\bsnm{Kuhn}, \binits{L.}},
\bauthor{\bsnm{Gal}, \binits{Y.}},
\bauthor{\bsnm{Farquhar}, \binits{S.}}:
\bctitle{Semantic uncertainty: Linguistic invariances for uncertainty
  estimation in natural language generation}.
In: \bbtitle{International Conference on Learning Representations (ICLR)}
(\byear{2023})
\end{bchapter}
\endbibitem

\bibitem[\protect\citeauthoryear{Joshi et~al.}{2017}]{joshi2017triviaqa}
\begin{bchapter}
\bauthor{\bsnm{Joshi}, \binits{M.}},
\bauthor{\bsnm{Choi}, \binits{E.}},
\bauthor{\bsnm{Weld}, \binits{D.S.}},
\bauthor{\bsnm{Zettlemoyer}, \binits{L.}}:
\bctitle{{TriviaQA}: A large scale distantly supervised challenge dataset for
  reading comprehension}.
In: \bbtitle{Proceedings of the Annual Meeting of the Association for
  Computational Linguistics (ACL 2017)}
(\byear{2017})
\end{bchapter}
\endbibitem

\bibitem[\protect\citeauthoryear{{Qwen Team}}{2024}]{qwen25}
\begin{botherref}
\oauthor{\bsnm{{Qwen Team}}}:
{Qwen2.5} technical report.
arXiv preprint arXiv:2412.15115
(2024)
\end{botherref}
\endbibitem

\bibitem[\protect\citeauthoryear{{Qwen Team}}{2025}]{qwen3}
\begin{botherref}
\oauthor{\bsnm{{Qwen Team}}}:
{Qwen3} technical report.
arXiv preprint arXiv:2505.09388
(2025)
\end{botherref}
\endbibitem

\bibitem[\protect\citeauthoryear{Grattafiori et~al.}{2024}]{llama3}
\begin{botherref}
\oauthor{\bsnm{Grattafiori}, \binits{A.}}, et al.:
The {Llama} 3 herd of models.
arXiv preprint arXiv:2407.21783
(2024)
\end{botherref}
\endbibitem

\bibitem[\protect\citeauthoryear{Jiang et~al.}{2023}]{mistral7b}
\begin{botherref}
\oauthor{\bsnm{Jiang}, \binits{A.Q.}}, et al.:
{Mistral} {7B}.
arXiv preprint arXiv:2310.06825
(2023)
\end{botherref}
\endbibitem

\bibitem[\protect\citeauthoryear{Abdin et~al.}{2024}]{phi4}
\begin{botherref}
\oauthor{\bsnm{Abdin}, \binits{M.}}, et al.:
{Phi-4} technical report.
arXiv preprint arXiv:2412.08905
(2024)
\end{botherref}
\endbibitem

\bibitem[\protect\citeauthoryear{Kwiatkowski et~al.}{2019}]{kwiatkowski2019nq}
\begin{barticle}
\bauthor{\bsnm{Kwiatkowski}, \binits{T.}},
\bauthor{\bsnm{Palomaki}, \binits{J.}},
\bauthor{\bsnm{Redfield}, \binits{O.}},
\bauthor{\bsnm{Collins}, \binits{M.}},
\bauthor{\bsnm{Parikh}, \binits{A.}},
\bauthor{\bsnm{Alberti}, \binits{C.}},
\bauthor{\bsnm{Epstein}, \binits{D.}},
\bauthor{\bsnm{Polosukhin}, \binits{I.}},
\bauthor{\bsnm{Devlin}, \binits{J.}},
\bauthor{\bsnm{Lee}, \binits{K.}}, \betal:
\batitle{Natural questions: A benchmark for question answering research}.
\bjtitle{Transactions of the Association for Computational Linguistics}
\bvolume{7},
\bfpage{452}--\blpage{466}
(\byear{2019})
\end{barticle}
\endbibitem

\bibitem[\protect\citeauthoryear{Yang et~al.}{2018}]{yang2018hotpot}
\begin{bchapter}
\bauthor{\bsnm{Yang}, \binits{Z.}},
\bauthor{\bsnm{Qi}, \binits{P.}},
\bauthor{\bsnm{Zhang}, \binits{S.}},
\bauthor{\bsnm{Bengio}, \binits{Y.}},
\bauthor{\bsnm{Cohen}, \binits{W.W.}},
\bauthor{\bsnm{Salakhutdinov}, \binits{R.}},
\bauthor{\bsnm{Manning}, \binits{C.D.}}:
\bctitle{{HotpotQA}: A dataset for diverse, explainable multi-hop question
  answering}.
In: \bbtitle{Proceedings of the 2018 Conference on Empirical Methods in Natural
  Language Processing (EMNLP)}
(\byear{2018})
\end{bchapter}
\endbibitem

\bibitem[\protect\citeauthoryear{Lin et~al.}{2022}]{lin2022truthfulqa}
\begin{bchapter}
\bauthor{\bsnm{Lin}, \binits{S.}},
\bauthor{\bsnm{Hilton}, \binits{J.}},
\bauthor{\bsnm{Evans}, \binits{O.}}:
\bctitle{{TruthfulQA}: Measuring how models mimic human falsehoods}.
In: \bbtitle{Proceedings of the Annual Meeting of the Association for
  Computational Linguistics (ACL 2022)}
(\byear{2022})
\end{bchapter}
\endbibitem

\bibitem[\protect\citeauthoryear{Berant et~al.}{2013}]{berant2013webq}
\begin{bchapter}
\bauthor{\bsnm{Berant}, \binits{J.}},
\bauthor{\bsnm{Chou}, \binits{A.}},
\bauthor{\bsnm{Frostig}, \binits{R.}},
\bauthor{\bsnm{Liang}, \binits{P.}}:
\bctitle{Semantic parsing on {Freebase} from question-answer pairs}.
In: \bbtitle{Proceedings of the 2013 Conference on Empirical Methods in Natural
  Language Processing (EMNLP)}
(\byear{2013})
\end{bchapter}
\endbibitem

\bibitem[\protect\citeauthoryear{Cobbe et~al.}{2021}]{cobbe2021gsm8k}
\begin{botherref}
\oauthor{\bsnm{Cobbe}, \binits{K.}},
\oauthor{\bsnm{Kosaraju}, \binits{V.}},
\oauthor{\bsnm{Bavarian}, \binits{M.}},
\oauthor{\bsnm{Chen}, \binits{M.}},
\oauthor{\bsnm{Jun}, \binits{H.}},
\oauthor{\bsnm{Kaiser}, \binits{L.}},
\oauthor{\bsnm{Plappert}, \binits{M.}},
\oauthor{\bsnm{Tworek}, \binits{J.}},
\oauthor{\bsnm{Hilton}, \binits{J.}},
\oauthor{\bsnm{Nakano}, \binits{R.}}, et al.:
Training verifiers to solve math word problems.
arXiv preprint arXiv:2110.14168
(2021)
\end{botherref}
\endbibitem

\bibitem[\protect\citeauthoryear{Tran}{2026a}]{prereg-pilot}
\begin{botherref}
\oauthor{\bsnm{Tran}, \binits{D.V.}}:
Pilot Preregistration: Multi-turn Confidence Geometry.
Frozen 2026-09-19 09:43 UTC, before generation. H2 failed; H1 inconclusive;
  locked verdict CONTINUE. Supplementary file PREREG.md (Online Resource 2).
(2026)
\end{botherref}
\endbibitem

\bibitem[\protect\citeauthoryear{Tran}{2026b}]{prereg-t3rep}
\begin{botherref}
\oauthor{\bsnm{Tran}, \binits{D.V.}}:
Preregistration: {T3} Replication of Stay-Wrong Persistence.
Frozen before generation, with two pre-generation amendments. Supplementary
  file PREREG\_T3REP.md (Online Resource 2).
(2026)
\end{botherref}
\endbibitem

\bibitem[\protect\citeauthoryear{Tran}{2026c}]{prereg-ext}
\begin{botherref}
\oauthor{\bsnm{Tran}, \binits{D.V.}}:
Preregistration: Judge Rescoring, Larger {T3} Samples, Retrieval with Real
  Evidence, and a {32B} Model.
Frozen before generation and judging, with five dated post-hoc notes.
  Supplementary file PREREG\_TMLR.md (Online Resource 2).
(2026)
\end{botherref}
\endbibitem

\bibitem[\protect\citeauthoryear{Robertson and
  Zaragoza}{2009}]{robertson2009bm25}
\begin{barticle}
\bauthor{\bsnm{Robertson}, \binits{S.}},
\bauthor{\bsnm{Zaragoza}, \binits{H.}}:
\batitle{The probabilistic relevance framework: {BM25} and beyond}.
\bjtitle{Foundations and Trends in Information Retrieval}
\bvolume{3}(\bissue{4}),
\bfpage{333}--\blpage{389}
(\byear{2009})
\end{barticle}
\endbibitem

\end{thebibliography}

\end{document}